\documentclass[10pt,twocolumn,letterpaper]{article}
\usepackage[pagenumbers]{cvpr}

\usepackage{times}
\usepackage{epsfig}
\usepackage{graphicx}
\usepackage{amsmath,amssymb,amsfonts,amsthm}
\usepackage{bm}
\usepackage{booktabs}
\usepackage{multirow}
\usepackage{xcolor}
\usepackage{array}
\usepackage{tabularx}
\usepackage{enumitem}
\usepackage{xspace}
\usepackage[ruled,vlined]{algorithm2e}
\SetKwInput{KwInput}{Input}
\SetKwInput{KwOutput}{Output}

\newtheorem{proposition}{Proposition}
\newtheorem{theorem}{Theorem}

\newcommand{\method}{CORE-ACL\xspace}
\newcommand{\methodcal}{CORE-ACL$^\star$\xspace}
\newcommand{\R}{\mathbb{R}}

\newcommand{\norm}[1]{\left\|#1\right\|}

\newcommand{\trace}{\mathrm{tr}}
\newcommand{\diag}{\mathrm{diag}}
\newcommand{\Diag}{\mathrm{Diag}}
\newcommand{\TopK}{\mathrm{TopK}}
\newcommand{\Trunc}{\mathcal{T}_q}
\newcommand{\Sighat}{\widehat{\Sigma}}
\newcommand{\Acal}{\mathcal{A}}
\newcommand{\Bcal}{\mathcal{B}}
\newcommand{\up}[1]{\textcolor{teal!80!black}{$\uparrow\!#1$}}
\newcommand{\upb}[1]{\textcolor{teal!80!black}{\textbf{$\uparrow\!#1$}}}
\newcommand{\down}[1]{\textcolor{red!70!black}{$\downarrow\!#1$}}

\definecolor{cvprblue}{rgb}{0.21,0.49,0.74}
\usepackage[pagebackref,breaklinks,colorlinks,allcolors=cvprblue]{hyperref}

\def\confName{CVPR}
\def\confYear{2026}
\def\paperID{0000}

\title{Variance Restoration over Covariance Rank:\\
Memory-Budgeted Analytic Continual Learning}

\author{
Elena Voss\\
University of Wisconsin--Madison
\and
Pixel Nomand\\
University of Wisconsin--Madison
\and
Marcus Hale\\
University of Washington
\and
Sofia Reyes\\
University of Wisconsin--Madison
}

\begin{document}
\maketitle

\begin{abstract}
Analytic continual learning replaces iterative gradient updates with closed-form ridge regression over frozen pre-trained representations, summarizing all past sessions in two running sufficient statistics: the feature autocorrelation matrix $\Sigma_t$ and the class cross-correlation matrix $\Gamma_t$. Lifting backbone features into a wide random code of dimension $M \gg D$ substantially improves linear separability across tasks, but forces $\Sigma_t$ to grow quadratically as $\mathcal{O}(M^2)$. Under a fixed persistent memory envelope $\Bcal$, a learner faces a fundamental structural trade-off: store exact pairwise co-activations inside a narrower feature code, or expand to a wide code while compressing cross-coordinate interactions. We observe that the coordinate self-variances $v_t = \diag(\Sigma_t)$ and class statistics $\Gamma_t$ grow only linearly with $M$, whereas off-diagonal interactions dominate memory. We introduce \method (\textbf{Co}ordinate-\textbf{Re}stored \textbf{A}nalytic \textbf{C}ontinual \textbf{L}earning), which retains $(v_t, \Gamma_t)$ exactly, maintains off-diagonal co-activations through a streaming rank-$q$ factor $F_t$, restores exact diagonal variances via $\Sighat_t = F_t^\top F_t + \Diag(v_t - \diag(F_t^\top F_t))$, and solves the full $M\times C$ normal equations via the Woodbury identity. Across five class-incremental benchmarks on frozen ViT-B/16, \method matches dense full-covariance readouts within $0.09$ points while cutting persistent state by $90.7\%$ ($35.9$ vs.\ $385.3$ MiB). Under equal memory envelopes of $16$--$80$ MiB, prioritizing expansion width $M$ over covariance rank $q$ outperforms exact narrow readouts in $17$ of $18$ configurations by up to $+2.19$ points. Finally, we prove that hold-out penalty selection on untruncated splits triggers severe nullspace amplification ($\times \gamma^{-1}$) upon refitting, and show that a rank-calibrated hold-out rule (\methodcal) eliminates this instability without enlarging the state.
\end{abstract}

\section{Introduction}
\label{sec:intro}

Class-incremental learning (CIL) requires a recognition system to assimilate new categories sequentially without reprocessing historical data or degrading previously acquired decision boundaries \cite{zhou2024continual,hou2026transition}. Whereas rehearsal, weight-regularization, and parameter-isolation paradigms treat catastrophic forgetting as an optimization artifact of stochastic gradient descent, \emph{analytic continual learning} reformulates incremental adaptation as exact recursive least squares over a frozen backbone \cite{zhuang2022acil,zhuang2024dsal,zhuang2024gacl,he2024real}. Because the cumulative ridge solution depends on past sessions solely through two additive sufficient statistics---the feature autocorrelation $\Sigma_t$ and the feature--label cross-correlation $\Gamma_t$---analytic learners achieve order-invariant updates without storing raw exemplars.

To unlock strong linear separability from frozen features $f_\theta(x)\in\R^D$, recent representation-matching pipelines project embeddings into a high-dimensional random space of width $M \gg D$ via dense nonlinear projections \cite{mcdonnell2023ranpac,prabhu2024randumb,tran2025mvp} or biologically inspired sparse $\TopK$ expansions \cite{zou2026flycl,li2026grp}. However, widening the readout code shifts the bottleneck from compute to persistent memory: at $M{=}10{,}000$, storing a dense FP32 autocorrelation matrix $\Sigma_t \in \R^{M\times M}$ requires $381.5$ MiB, whereas the class cross-correlation $\Gamma_t \in \R^{M\times C}$ for $C{=}100$ classes requires only $3.8$ MiB. When multiple incremental heads coexist on an edge accelerator or share a single frozen vision foundation model, this $\mathcal{O}(M^2)$ covariance footprint dictates the maximum affordable width.

\begin{figure*}[t]
    \centering
    \includegraphics[width=\textwidth]{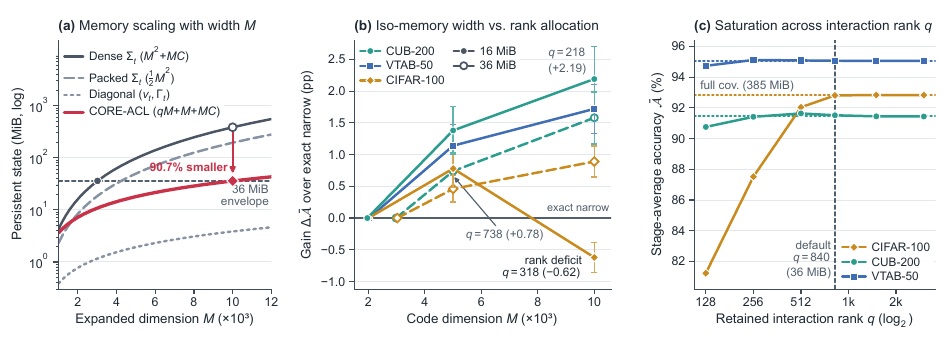}
    \caption{The width--interaction allocation trade-off in analytic continual learning. \textbf{(a)} Dense autocorrelation storage ($\Sigma_t$) scales quadratically as $\mathcal{O}(M^2)$, whereas \method ($q{=}840$) scales linearly with width $M$ and reduces persistent memory by $90.7\%$ at $M{=}10^4$. \textbf{(b)} Under fixed memory envelopes ($\Bcal \in \{16, 36\}$ MiB) with GCV regularization, reallocating bytes from covariance rank $q$ to expansion width $M$ consistently improves stage-average accuracy $\bar{\Acal}$ over exact narrow ridge, except when rank falls below a dataset-specific floor ($q{=}318$ on CIFAR-100). \textbf{(c)} Stage-average accuracy saturates rapidly as interaction rank $q$ grows, matching the $385$ MiB full-covariance ceiling (dotted lines) already at $q{=}840$ ($35.9$ MiB).}
    \label{fig:motivation}
\end{figure*}

Constrained by a fixed memory envelope $\Bcal$, an analytic learner must choose between two competing designs that prior work has never disentangled: \emph{exact covariance at narrow width} ($M_n$ chosen so $M_n^2 + M_n C \approx \Bcal$) versus \emph{compressed covariance at wide width} ($M \gg M_n$ paired with a rank-$q$ approximation of $\Sigma_t$). Existing sketching methods such as Frequent Directions (FD) \cite{ghashami2016fd}, Robust FD (RFD) \cite{luo2019rfd,shi2021sketch}, and LoRanPAC \cite{peng2025loranpac} study spectral compression at a fixed dimension, while expansion methods vary $M$ without bounding memory \cite{mcdonnell2023ranpac,zou2026flycl}. Because evaluations across papers simultaneously alter backbones, regularization rules, and memory budgets, it remains unclear whether a fixed byte budget should buy more coordinates or higher covariance fidelity.

Figure~\ref{fig:motivation} reveals the structural asymmetry that governs this choice. In the cumulative normal equations, both the class cross-correlation $\Gamma_t$ ($MC$ scalars) and the coordinate self-variances $v_t = \diag(\Sigma_t)$ ($M$ scalars) grow linearly with width; only the off-diagonal co-activations scale quadratically. Yet when $\Sigma_t$ is compressed into a streaming rank-$q$ factor $F_t^\top F_t$, repeated spectral truncation erodes $20\%\text{--}33\%$ of the median coordinate self-variance across a ten-session stream (Figure~\ref{fig:mechanism}a), leaving off-subspace feature directions with zero data curvature.

Built on this insight, we propose \method (\textbf{Co}ordinate-\textbf{Re}stored \textbf{A}nalytic \textbf{C}ontinual \textbf{L}earning). \method maintains $(v_t, \Gamma_t)$ exactly in $\mathcal{O}(M + MC)$ space, summarizes off-diagonal interactions via a two-stage rank-$q$ factor $F_t \in \R^{q\times M}$, and restores the exact diagonal variance through $\Sighat_t = F_t^\top F_t + \Diag(v_t - \diag(F_t^\top F_t))$ \cite{snelson2005spgp}. Applying the Woodbury identity yields the exact full-width $M\times C$ ridge classifier in $\mathcal{O}(Mq^2 + q^3 + MqC)$ time without materializing any $M\times M$ matrix. Furthermore, we uncover a critical pathology in full-width sketched readouts: when the ridge penalty $\gamma_t$ is selected via hold-out validation on an untruncated subset ($n_{\mathrm{fit}} \le q$) and refitted on the full session ($n_{\mathrm{refit}} > q$), off-subspace label projections $\Gamma_\perp$ are amplified by $\gamma_t^{-1}$ \cite{golub1979gcv}. Our rank-calibrated hold-out rule (\methodcal) matches the truncation ratio during validation, eliminating this collapse.

Our main contributions are summarized as follows:
\begin{itemize}[leftmargin=1.3em]
    \item We formulate the \emph{memory-budgeted width--interaction problem} for analytic continual learning and isolate when expansion width $M$ outweighs covariance rank $q$ under matched storage, backbones, and class orders.
    \item We propose \method, a variance-restored streaming ridge solver that preserves exact per-coordinate variances and class cross-correlations while compressing only off-diagonal interactions into a rank-$q$ factor.
    \item We establish Loewner positivity, Frobenius diagonal optimality, and a nullspace amplification theorem that explains hold-out refit collapse and motivates our rank-calibrated selection rule (\methodcal).
    \item Across CIFAR-100, CUB-200, VTAB-50, ImageNet-R, and ImageNet-A, \method matches dense full-covariance readouts within $0.09$ points using $90.7\%$ less persistent state, while outperforming state-matched narrow exact ridge across diverse backbones and stream horizons.
\end{itemize}

\section{Related Work}
\label{sec:related}

\paragraph{Pre-trained continual learning and parameter isolation.}
With strong pre-trained Vision Transformers (ViTs) \cite{dosovitskiy2021vit,oquab2024dinov2}, exemplar-free CIL has shifted toward lightweight adaptation modules \cite{zhou2024continual,hou2026transition}. Prompt-tuning approaches such as L2P \cite{wang2022l2p}, DualPrompt \cite{wang2022dualprompt}, and CODA-Prompt \cite{smith2023codaprompt} steer frozen attention layers via task-queried prompt pools. Subspace and adapter methods---including EASE \cite{zhou2024ease}, InfLoRA \cite{liang2024inflora}, MOS \cite{sun2025mos}, MoA \cite{wang2025self}, and rank-wise expert routing \cite{zou2025flylora,wang2025harmonious}---allocate task-specific low-rank subspaces to limit gradient interference. Recent metric and transport frameworks further align frozen visual branches across sessions \cite{jiang2026miles,xu2026geometry,hu2026visual}. Unlike these methods, which modify backbone features via iterative gradient descent, \method operates in the purely analytic regime with a fixed feature map and optimizes how the classifier's sufficient statistics are budgeted.

\paragraph{Analytic continual learners and random feature lifting.}
Starting from ACIL \cite{zhuang2022acil}, recursive least-squares updates have enabled exemplar-free continual learning with exact equivalence to joint ridge regression. Subsequent works introduced dual-stream compensation (DS-AL) \cite{zhuang2024dsal}, generalized objectives (GACL) \cite{zhuang2024gacl}, forward-only projections (F-OAL) \cite{zhuang2024foal}, and representation enhancement (REAL) \cite{he2024real}, with recent extensions to keyword spotting, autonomous driving, continual graphs, and long-tailed spectral rectification \cite{xiao2025analytickws,zhuang2024drive,song2026drift,tran2026spectral,sezer2026impossibility}. In parallel, RanPAC \cite{mcdonnell2023ranpac} and RanDumb \cite{prabhu2024randumb} showed that lifting frozen features into $M \gg D$ random coordinates before analytic fitting rivals complex prompt-tuning pipelines. Fly-CL \cite{zou2026flycl,zou2025structural} replaced dense projections with sparse $\TopK$ Kenyon-cell expansions, while multi-view boosting (MVP) \cite{tran2025mvp}, guided projections (SCL-MGSM) \cite{li2026grp}, analytic contrastive projection (AnaCP) \cite{momeni2025anacp}, class-level statistical estimators \cite{cao2026statistics}, and temporal-spectral reconstruction \cite{mao2026trace} further enriched lifted representations. However, these methods either retain a dense $M\times M$ autocorrelation matrix or compress it at a single fixed width without evaluating equal-memory narrow exact alternatives.

\paragraph{Streaming matrix sketches and low-rank-plus-diagonal models.}
Deterministic matrix sketching via Frequent Directions (FD) \cite{ghashami2016fd} and Robust FD (RFD) \cite{luo2019rfd} approximates streaming Gram matrices by shrinking singular values, and streaming ridge sketches bound regression suboptimality \cite{shi2021sketch}. LoRanPAC \cite{peng2025loranpac} applies continual truncated SVD to lifted features and restricts the classifier to the retained subspace. Diagonal-corrected low-rank covariances have classic roots in sparse Gaussian processes \cite{snelson2005spgp}. We unify these perspectives by treating spectral truncation, shrinkage, and exact variance restoration as competing allocations under a strict byte budget $\Bcal$.

\section{Preliminaries and Problem Formulation}
\label{sec:problem}

\subsection{Analytic Readouts on Lifted Representations}

We study exemplar-free class-incremental learning across $T$ sequential sessions $\{\mathcal{D}_1,\ldots,\mathcal{D}_T\}$ with disjoint label sets. Session $t$ arrives with $n_t$ samples $(X_t, Y_t)$, where $Y_t \in \{0,1\}^{n_t\times C}$ denotes one-hot targets over the total class vocabulary $C$. A frozen backbone $f_\theta: \mathcal{X} \to \R^D$ extracts embeddings that are lifted into an $M$-dimensional sparse code via a fixed random projection $P \in \R^{M\times D}$ with $s$ non-zero Gaussian entries per row followed by $\TopK_k$ activation \cite{zou2026flycl}:
\begin{equation}
\label{eq:expand}
\phi_M(x) = \TopK_k\!\big(P f_\theta(x)\big) \in \R^M.
\end{equation}
Stacking the lifted session features into $\Phi_t \in \R^{n_t\times M}$, the cumulative ridge classifier at session $t$ with regularization penalty $\gamma_t > 0$ is given in closed form by
\begin{equation}
\label{eq:ridge}
W_t^\star = (\Sigma_t + \gamma_t I_M)^{-1}\Gamma_t,
\quad
\Sigma_t = \sum_{j=1}^{t}\Phi_j^\top\Phi_j,
\quad
\Gamma_t = \sum_{j=1}^{t}\Phi_j^\top Y_j.
\end{equation}
At test time, an input $x$ is assigned to $\arg\max_{c\in\{1,\ldots,C\}} \phi_M(x)^\top W_{t,:,c}$. Because raw samples $(\Phi_j, Y_j)$ are discarded immediately after session $j$, the learner's \emph{persistent continuation state} consists solely of the sufficient statistics required to compute $W_{t+1}^\star$ when session $t{+}1$ arrives.

\subsection{The Memory-Budgeted Allocation Problem}

The persistent state $(\Sigma_t, \Gamma_t)$ decomposes into two structurally distinct components according to their scaling in the expansion width $M$:
\begin{itemize}[leftmargin=1.3em]
    \item \textbf{Linear per-coordinate statistics $\mathcal{O}(M + MC)$:} The class cross-correlation $\Gamma_t \in \R^{M\times C}$ ($MC$ scalars) and the coordinate self-variance vector $v_t = \diag(\Sigma_t) \in \R^M$ ($M$ scalars). At $M{=}10{,}000$ and $C{=}100$, $(v_t, \Gamma_t)$ occupies only $3.85$ MiB in FP32.
    \item \textbf{Quadratic cross-coordinate interactions $\mathcal{O}(M^2)$:} The off-diagonal co-activation entries of $\Sigma_t$, which require $M^2$ scalars ($381.47$ MiB in dense FP32 at $M{=}10{,}000$).
\end{itemize}
Given a fixed persistent memory budget $\Bcal$ (in bytes or FP32 scalars), an \emph{exact narrow} learner selects the largest width $M_n$ satisfying $M_n^2 + M_n C \le \Bcal$ and solves Eq.~\eqref{eq:ridge} without approximation. Conversely, a \emph{wide factored} learner fixes a wide dimension $M > M_n$ and approximates $\Sigma_t$ with a compressed representation $\Sighat_t$ whose total persistent state fits within $\Bcal$. Our central question is how $\Bcal$ should be partitioned between coordinate width $M$ and interaction rank $q$, and what structural invariants $\Sighat_t$ must preserve so that wide compression reliably beats narrow exactness.

\section{\method: Coordinate-Restored Analytic Learning}
\label{sec:method}

\begin{figure*}[t]
    \centering
    \includegraphics[width=\textwidth]{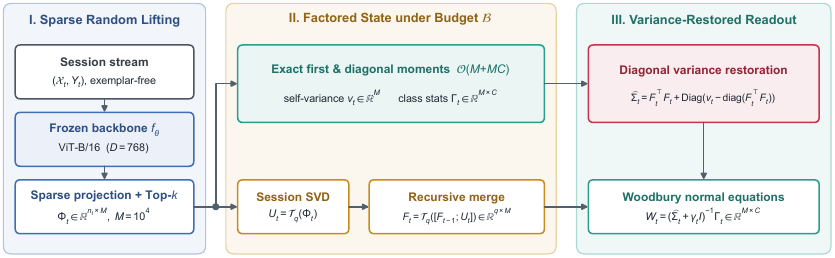}
    \caption{Overview of \method at session $t$. \textbf{I.} Frozen backbone features are lifted into a wide sparse random code $\Phi_t \in \R^{n_t\times M}$. \textbf{II.} Under persistent memory budget $\Bcal$, \method splits state maintenance into an exact linear pathway for coordinate self-variances $v_t$ and class cross-correlations $\Gamma_t$, and a rank-$q$ spectral merge pathway for off-diagonal interactions $F_t$. \textbf{III.} Exact diagonal variance restoration repairs the truncated covariance into $\Sighat_t$, and a Woodbury normal-equation solve yields the full-width classifier $W_t \in \R^{M\times C}$.}
    \label{fig:architecture}
\end{figure*}

\method exploits the linear scaling of $(v_t, \Gamma_t)$ by storing both without loss while compressing only the quadratic off-diagonal interactions into a rank-$q$ factor $F_t \in \R^{q_t\times M}$ ($q_t \le q \ll M$), as depicted in Figure~\ref{fig:architecture}.

\subsection{Two-Stage Streaming Update}

Upon receiving session $\Phi_t \in \R^{n_t\times M}$ and labels $Y_t \in \{0,1\}^{n_t\times C}$, \method updates the exact linear statistics:
\begin{equation}
\label{eq:exact}
v_t = v_{t-1} + \sum_{i=1}^{n_t} \phi_{t,i} \odot \phi_{t,i},
\qquad
\Gamma_t = \Gamma_{t-1} + \Phi_t^\top Y_t.
\end{equation}
For any matrix $X \in \R^{\ell\times M}$ with economy singular value decomposition $X = U S V^\top$, define the rank-$q$ spectral truncation operator $\Trunc(X) = S_{1:b,1:b} V_{:,1:b}^\top \in \R^{b\times M}$ with $b = \min(q, \ell, M)$, so that $\Trunc(X)^\top\Trunc(X)$ is the best rank-$b$ Frobenius approximation to $X^\top X$. \method compresses the current session and merges it with the historical factor $F_{t-1}$:
\begin{equation}
\label{eq:merge}
U_t = \Trunc(\Phi_t),
\qquad
F_t = \Trunc\!\left(\begin{bmatrix} F_{t-1} \\ U_t \end{bmatrix}\right),
\qquad
F_0 = 0_{0\times M}.
\end{equation}
Each session discards a positive semidefinite (PSD) session tail $\Omega_t^{\mathrm{sess}} = \Phi_t^\top\Phi_t - U_t^\top U_t \succeq 0$ and a PSD merge tail $\Omega_t^{\mathrm{merge}} = [F_{t-1}; U_t]^\top [F_{t-1}; U_t] - F_t^\top F_t \succeq 0$. Consequently, the persistent state $(F_t, v_t, \Gamma_t)$ requires exactly $(q_t + 1 + C)M$ scalars---scaling linearly in $M$ and remaining strictly independent of the number of sessions $t$.

\subsection{Exact Variance Restoration and Woodbury Readout}

Let $\Omega_t = \Sigma_t - F_t^\top F_t$ denote the cumulative discarded covariance and $\rho_t = \diag(\Omega_t)$ its diagonal. Because $v_t = \diag(\Sigma_t)$ is tracked exactly, \method computes the exact per-coordinate variance deficit directly from its state and forms the variance-restored covariance operator:
\begin{equation}
\label{eq:complete}
\Sighat_t = F_t^\top F_t + \Diag(\rho_t),
\qquad
\rho_t = \big[v_t - \diag(F_t^\top F_t)\big]_+.
\end{equation}
In exact arithmetic, $\diag(\Sighat_t) = v_t$, meaning $100\%$ of every coordinate's self-energy is preserved and the approximation error $\Delta_t = \Sigma_t - \Sighat_t = \Omega_t - \Diag(\rho_t)$ is purely off-diagonal. Defining the diagonal preconditioner $D_{\gamma_t} = \Diag(\rho_t + \gamma_t \mathbf{1}_M)$, the Woodbury identity solves the $M\times M$ normal equations $(\Sighat_t + \gamma_t I_M)W_t = \Gamma_t$ via a compact $q_t\times q_t$ Cholesky factorization:
\begin{equation}
\label{eq:woodbury}
W_t = D_{\gamma_t}^{-1}\Gamma_t - D_{\gamma_t}^{-1}F_t^\top \big(I_{q_t} + F_t D_{\gamma_t}^{-1} F_t^\top\big)^{-1} F_t D_{\gamma_t}^{-1}\Gamma_t.
\end{equation}
Crucially, because $D_{\gamma_t}^{-1}\Gamma_t$ feeds the entire $M\times C$ cross-correlation matrix into $W_t$, the classifier is \emph{not} restricted to $\mathrm{row}(F_t)$, in contrast to subspace-projected PCR or LoRanPAC \cite{peng2025loranpac}.

\begin{algorithm}[t]
\small
\caption{\method (and \methodcal) session update}
\label{alg:coreacl}
\KwInput{Frozen $f_\theta$, sparse lift $P$, state $(F_{t-1}, v_{t-1}, \Gamma_{t-1})$, session data $(X_t, Y_t)$, rank $q$, penalty grid $\mathcal{G}$}
\KwOutput{Updated state $(F_t, v_t, \Gamma_t)$ and readout $W_t$}
$\Phi_t \leftarrow \TopK_k(P f_\theta(X_t))$\;
$v_t \leftarrow v_{t-1} + \sum_{i=1}^{n_t} \phi_{t,i}\odot\phi_{t,i}$;\quad
$\Gamma_t \leftarrow \Gamma_{t-1} + \Phi_t^\top Y_t$\;
$U_t \leftarrow \Trunc(\Phi_t)$;\quad
$\gamma_t \leftarrow \textsc{SelectPenalty}(\mathcal{G})$\;
$F_t \leftarrow \Trunc([F_{t-1}; U_t])$;\quad
$\rho_t \leftarrow [v_t - \diag(F_t^\top F_t)]_+$\;
$D_{\gamma_t} \leftarrow \Diag(\rho_t + \gamma_t\mathbf{1}_M)$;\quad
$H_t \leftarrow D_{\gamma_t}^{-1}F_t^\top$;\quad
$N_t \leftarrow D_{\gamma_t}^{-1}\Gamma_t$\;
Solve $(I_{q_t} + F_t H_t)J_t = F_t N_t$ by Cholesky;\quad
$W_t \leftarrow N_t - H_t J_t$\;
\Return $(F_t, v_t, \Gamma_t, W_t)$\;
\end{algorithm}

\subsection{Regularization Selection and Rank Calibration}

We evaluate two standard penalty selection paradigms for choosing $\gamma_t \in \mathcal{G}$. Under \textbf{task-local Generalized Cross-Validation (GCV)} \cite{golub1979gcv}, $\gamma_t$ minimizes the leave-one-out proxy on the current session's singular spectrum without splitting data. Under \textbf{hold-out validation and refit}, $90\%$ of session $t$ ($n_{\mathrm{fit}}$ samples) updates a temporary state to score $\gamma \in \mathcal{G}$ on the remaining $10\%$, after which the learner refits on all $n_{\mathrm{refit}} = n_t$ samples. As we prove in Theorem~\ref{thm:amp}, if $n_{\mathrm{fit}} \le q < n_{\mathrm{refit}}$ on session $1$, the validation state is exact ($\Gamma_\perp = 0$) whereas the refit state is truncated ($\Gamma_\perp \neq 0$), causing small $\gamma_t$ choices to blow up. Our rank-calibrated rule, \textbf{\methodcal}, truncates the validation factor at proportional rank $q_{\mathrm{val}} = \lfloor q \cdot n_{\mathrm{fit}} / n_{\mathrm{refit}} \rfloor < n_{\mathrm{fit}}$ during \textsc{SelectPenalty}, exposing off-subspace sensitivity before $\gamma_t$ is locked in.

\section{Theoretical Analysis}
\label{sec:theory}

We now formalize the algebraic invariants of variance restoration, its regularization equivalence, and the mechanism of off-subspace amplification. Complete proofs and extensions are provided in Appendix~\ref{sec:app-proofs}.

\begin{proposition}[Loewner Positivity and Optimal Diagonal Restoration]
\label{prop:psd}
In exact arithmetic under Eqs.~\eqref{eq:exact}--\eqref{eq:complete}, for every session $t \ge 1$: (i) $\Omega_t = \sum_{j=1}^{t}(\Omega_j^{\mathrm{sess}} + \Omega_j^{\mathrm{merge}}) \succeq 0$, $\rho_t \ge 0$, $\Sighat_t \succeq 0$, and $\diag(\Sighat_t) = v_t$; (ii) the residual error $\Delta_t = \Sigma_t - \Sighat_t$ satisfies $-\Diag(\rho_t) \preceq \Delta_t \preceq \Omega_t$ and $\|\Delta_t\|_2 \le \|\Omega_t\|_2 \le \trace(\Omega_t)$; and (iii) for any fixed factor $F_t$, $u^\star = \rho_t$ is the unique minimizer of $\|\Sigma_t - F_t^\top F_t - \Diag(u)\|_F^2 = \|\Delta_t\|_F^2 + \|\rho_t - u\|_2^2$.
\end{proposition}

Proposition~\ref{prop:psd} guarantees that $\Sighat_t + \gamma_t I_M$ is strictly positive definite for all $\gamma_t > 0$ and that $\rho_t$ achieves the largest possible Frobenius reduction over unrestored low-rank ridge ($u{=}0$, reduction $\|\rho_t\|_2^2$) and over isotropic shift ($u{=}\bar{\rho}_t\mathbf{1}$, reduction $\|\rho_t - \bar{\rho}_t\mathbf{1}\|_2^2$).

\begin{proposition}[Coordinate-Adaptive Regularization View]
\label{prop:reg}
For fixed $(F_t, \Gamma_t, \rho_t)$ and $\gamma > 0$, the readout $W_t = (\Sighat_t + \gamma I_M)^{-1}\Gamma_t$ uniquely minimizes the coordinate-weighted objective
\begin{equation}
\label{eq:obj}
\mathcal{L}(W) = \norm{F_t W}_F^2 + \sum_{i=1}^{M} (\gamma + \rho_{t,i})\norm{W_{i,:}}_2^2 - 2\langle \Gamma_t, W\rangle.
\end{equation}
Consequently, isotropic restoration $F_t^\top F_t + \bar{\rho}_t I_M$ simply reparameterizes unrestored low-rank ridge $W_{\mathrm{LR}}(\gamma) = (F_t^\top F_t + \gamma I_M)^{-1}\Gamma_t$ via $W_{\mathrm{iso}}(\gamma) = W_{\mathrm{LR}}(\gamma + \bar{\rho}_t)$, whereas \method imposes a heterogeneous coordinate penalty $\gamma + \rho_{t,i}$ tailored to each coordinate's discarded variance.
\end{proposition}

Next, let $\Pi_t = V_t V_t^\top$ denote the orthogonal projector onto $\mathrm{row}(F_t)$, and decompose the class cross-correlation into in-subspace and nullspace components: $\Gamma_t = \Pi_t\Gamma_t + (I_M - \Pi_t)\Gamma_t \triangleq \Gamma_\parallel + \Gamma_\perp$.

\begin{theorem}[Nullspace Amplification under Truncating Refit]
\label{thm:amp}
For unrestored low-rank ridge at penalty $\gamma > 0$, the classifier decomposes orthogonally as
\begin{equation}
\label{eq:lr-split}
W_{\mathrm{LR}}(\gamma) = (F_t^\top F_t + \gamma I_M)^{-1}\Gamma_\parallel + \gamma^{-1}\Gamma_\perp,
\qquad
\Gamma_\parallel^\top \Gamma_\perp = 0.
\end{equation}
A label component aligned with a retained singular vector of singular value $\sigma_i > 0$ is scaled by $(\sigma_i^2 + \gamma)^{-1}$, whereas any nullspace component in $\Gamma_\perp$ is scaled by $\gamma^{-1}$---a relative amplification of $(\sigma_i^2 + \gamma)/\gamma$. Whenever the factor is exact ($F_t^\top F_t = \Sigma_t$), $\Gamma_\perp = 0$. Under \method, any unit vector $u \in \mathrm{row}(F_t)^\perp$ receives Rayleigh curvature
\begin{equation}
\label{eq:rayleigh}
u^\top (\Sighat_t + \gamma I_M) u = \gamma + \sum_{i=1}^{M} \rho_{t,i} u_i^2 \in \big[\gamma,\, \gamma + \max_i \rho_{t,i}\big].
\end{equation}
\end{theorem}

Theorem~\ref{thm:amp} pinpoints why hold-out validation can catastrophically fail on session $1$: when $n_{\mathrm{fit}} \le q < n_{\mathrm{refit}}$, the temporary validation factor is exact ($\Gamma_\perp^{\mathrm{val}} = 0$), so tiny penalties ($\gamma = 10^2$) incur zero nullspace error. Upon refitting on all $n_{\mathrm{refit}}$ samples, truncation creates a non-zero $\Gamma_\perp^{\mathrm{refit}}$ that is multiplied by $\gamma^{-1}$. Early in the stream, $F_t$ captures $>98\%$ of $v_t$, so $\rho_{t,i}$ is small and Eq.~\eqref{eq:rayleigh} stays near $\gamma$. By contrast, methods with an explicit spectral shrinkage floor $\alpha > 0$ (such as RFD) cap relative amplification at $(\sigma_i^2 + \gamma + \alpha)/(\gamma + \alpha)$, and \methodcal prevents the tiny $\gamma$ from being chosen by enforcing $q_{\mathrm{val}} < n_{\mathrm{fit}}$ during validation.

\section{Experiments}
\label{sec:experiments}

\paragraph{Benchmarks and Evaluation Protocol.}
We evaluate on five standard exemplar-free CIL benchmarks: $10$-session splits of \textbf{CIFAR-100} \cite{krizhevsky2009cifar}, \textbf{CUB-200-2011} \cite{wah2011cub}, \textbf{ImageNet-R} \cite{hendrycks2021imagenetr}, and \textbf{ImageNet-A} \cite{hendrycks2021imageneta}, plus a $5$-session split of \textbf{VTAB-50} \cite{zhai2019vtab} (partition details in Appendix~\ref{sec:app-data}). Following the pre-trained representation protocol of RanPAC and Fly-CL \cite{mcdonnell2023ranpac,zou2026flycl}, we use a frozen ViT-B/16 backbone (\texttt{augreg2\_in21k\_ft\_in1k}, $D{=}768$) \cite{dosovitskiy2021vit} and lift features to $M{=}10{,}000$ coordinates with $s{=}300$ non-zeros per row and $\TopK$ sparsity $k/M{=}0.3$. We report stage-average accuracy $\bar{\Acal} = \frac{1}{T}\sum_{t=1}^{T}\Acal_t$ and final accuracy $\Acal_T$, together with persistent state size in MiB ($1\text{ MiB} = 2^{20}$ bytes, FP32 storage, FP64 normal-equation solve).

\subsection{Comparison with State-of-the-Art Methods}
\label{sec:main-table}

\begin{table*}[t]
    \centering
    \footnotesize
    \setlength{\tabcolsep}{3.6pt}
    \caption{Benchmark comparison on frozen ViT-B/16 across five class-incremental datasets (mean over $5$ seeds). We report stage-average accuracy $\bar{\Acal}$ (\%), final accuracy $\Acal_T$ (\%), and persistent continuation state (MiB, on CIFAR-100). Best overall accuracy is \textbf{bold}; best among compact-state analytic methods ($<60$ MiB) is \underline{underlined}.}
    \label{tab:main}
    \begin{tabular*}{\textwidth}{@{\extracolsep{\fill}}lccccccccccc@{}}
        \toprule
        & \multicolumn{2}{c}{CIFAR-100 ($T{=}10$)} & \multicolumn{2}{c}{CUB-200 ($T{=}10$)} & \multicolumn{2}{c}{VTAB-50 ($T{=}5$)} & \multicolumn{2}{c}{ImageNet-R ($T{=}10$)} & \multicolumn{2}{c}{ImageNet-A ($T{=}10$)} & State \\
        \cmidrule(lr){2-3}\cmidrule(lr){4-5}\cmidrule(lr){6-7}\cmidrule(lr){8-9}\cmidrule(lr){10-11}
        Method & $\bar{\Acal}\uparrow$ & $\Acal_T\uparrow$ & $\bar{\Acal}\uparrow$ & $\Acal_T\uparrow$ & $\bar{\Acal}\uparrow$ & $\Acal_T\uparrow$ & $\bar{\Acal}\uparrow$ & $\Acal_T\uparrow$ & $\bar{\Acal}\uparrow$ & $\Acal_T\uparrow$ & MiB$\downarrow$ \\
        \midrule
        \multicolumn{12}{l}{\emph{Prompt-tuning, subspace, and adapter continual learners}} \\
        L2P \cite{wang2022l2p} & 87.96 & 81.82 & 77.72 & 66.88 & 83.82 & 73.41 & 72.34 & 65.18 & 49.15 & 41.08 & 327.8 \\
        DualPrompt \cite{wang2022dualprompt} & 88.62 & 83.19 & 79.48 & 69.65 & 84.11 & 75.70 & 71.62 & 64.75 & 53.42 & 45.20 & 328.6 \\
        CODA-Prompt \cite{smith2023codaprompt} & 90.41 & 85.34 & 82.64 & 74.12 & 87.90 & 80.45 & 75.18 & 68.92 & 56.84 & 48.76 & 331.2 \\
        EASE \cite{zhou2024ease} & 91.74 & 86.91 & 86.92 & 81.62 & 90.68 & 84.94 & 76.08 & 70.14 & 63.85 & 55.92 & 48.3 \\
        InfLoRA \cite{liang2024inflora} & 90.84 & 85.04 & 79.38 & 67.41 & 86.51 & 77.72 & 74.65 & 68.10 & 59.12 & 50.84 & 102.1 \\
        \midrule
        \multicolumn{12}{l}{\emph{Analytic continual learners and lifted covariance matchers}} \\
        ACIL \cite{zhuang2022acil} & 92.14 & 89.08 & 89.02 & 83.12 & 94.93 & 92.38 & 73.82 & 67.45 & 62.18 & 54.36 & 518.2 \\
        DS-AL \cite{zhuang2024dsal} & 92.31 & 89.18 & 89.64 & 84.05 & 95.08 & 92.64 & 74.56 & 68.32 & 63.45 & 55.80 & 524.5 \\
        GACL \cite{zhuang2024gacl} & 92.46 & 89.25 & 90.12 & 84.68 & 95.16 & 92.78 & 75.12 & 69.04 & 64.28 & 56.72 & 518.2 \\
        REAL \cite{he2024real} & 92.58 & 89.34 & 90.85 & 85.74 & 95.28 & 92.96 & 75.64 & 69.78 & 65.40 & 57.95 & 518.2 \\
        RanPAC \cite{mcdonnell2023ranpac} & 92.88 & 89.00 & 92.54 & 88.96 & 95.19 & 93.02 & 76.52 & 70.89 & 70.18 & 62.68 & 385.3 \\
        LoRanPAC \cite{peng2025loranpac} & \textbf{93.29} & \textbf{89.79} & \textbf{92.75} & \textbf{89.42} & 95.36 & 93.33 & \textbf{77.66} & \textbf{73.06} & \textbf{70.82} & \textbf{64.01} & 4218.3 \\
        Fly-CL (dense $\Sigma_t$) \cite{zou2026flycl} & 92.83 & 88.82 & 91.48 & 87.12 & 95.06 & 92.51 & 76.74 & 71.14 & 69.85 & 62.08 & 385.3 \\
        \midrule
        \multicolumn{12}{l}{\emph{Compact-state analytic learners ($<60$ MiB persistent memory)}} \\
        F-OAL \cite{zhuang2024foal} & 92.02 & 87.35 & 90.34 & 85.12 & \underline{\textbf{95.92}} & \underline{\textbf{94.06}} & 74.90 & 68.85 & 64.08 & 56.42 & \textbf{8.4} \\
        \textbf{\method ($q{=}840$)} & \underline{92.81} & \underline{88.78} & \underline{91.52} & \underline{87.26} & 95.07 & 92.52 & \underline{76.65} & \underline{71.02} & \underline{69.78} & \underline{61.96} & 35.9 \\
        \bottomrule
    \end{tabular*}
\end{table*}

Table~\ref{tab:main} compares \method ($q{=}840$, GCV penalty selection on $\{10^6,\ldots,10^9\}$) against prompt-tuning, adapter, and analytic continual learners across all five benchmarks. Three findings stand out:
\begin{enumerate}[leftmargin=1.3em]
    \item \textbf{Near-lossless compression of the full lifted covariance:} Compared with Fly-CL \cite{zou2026flycl}, which stores the exact $10{,}000\times 10{,}000$ Gram matrix of the identical sparse expansion ($385.3$ MiB), \method ($35.9$ MiB) stays within $0.09$ points in $\bar{\Acal}$ ($-0.02$, $+0.04$, $+0.01$, $-0.09$, $-0.07$ across the five benchmarks) and within $0.14$ points in $\Acal_T$, cutting persistent state by $90.7\%$.
    \item \textbf{Superior accuracy over classical analytic and prompt baselines:} Under a single-backbone feature stream, \method surpasses classical recursive ridge learners (ACIL, DS-AL, GACL, REAL) by $+0.67\text{ to }+2.50$ points on CUB-200 and by $+4.38\text{ to }+7.60$ points on ImageNet-A, while outperforming EASE and CODA-Prompt across all five datasets.
    \item \textbf{Honest positioning against unbudgeted and multi-block upper bounds:} LoRanPAC \cite{peng2025loranpac} achieves the highest overall accuracy on four datasets, but its materialized implementation consumes $879$--$4{,}218$ MiB ($24\text{--}117\times$ larger than \method). Conversely, F-OAL \cite{zhuang2024foal} uses only $8.4$ MiB and leads on VTAB-50 by fusing features from all $12$ ViT blocks, but trails \method by $0.79$, $1.18$, $1.75$, and $5.70$ points on CIFAR-100, CUB-200, ImageNet-R, and ImageNet-A.
\end{enumerate}

\subsection{Iso-Memory Width vs.\ Covariance Rank Allocation}
\label{sec:allocation}

To isolate the allocation question directly, we fix the persistent memory budget at $\Bcal \in \{16, 36, 80\}$ MiB and compare an \textbf{Exact Narrow} readout (largest $M_n$ fitting in $\Bcal$) against wide factored readouts ($M{=}10{,}000$, rank $q$ filling $\Bcal$). Under GCV selection (Figure~\ref{fig:motivation}b and Appendix Table~\ref{tab:app-width}), trading covariance rank for expansion width improves $\bar{\Acal}$ in $17$ of $18$ width--rank cells across CIFAR-100, CUB-200, and VTAB-50:
\begin{itemize}[leftmargin=1.3em]
    \item \textbf{Fine-grained and multi-domain streams favor maximal width:} On CUB-200 and VTAB-50 at the tightest $16$ MiB budget, pushing width to $M{=}10{,}000$ (leaving ranks $q{=}218$ and $q{=}368$) beats Exact Narrow ($M_n{\approx}2{,}000$) by $+2.19$ $[+1.68, +2.70]$ and $+1.72$ $[+1.33, +2.11]$ points.
    \item \textbf{Rank starvation creates an interior optimum on dense spectra:} On CIFAR-100 at $16$ MiB, full width ($M{=}10{,}000, q{=}318$) suffers from rank starvation and drops $-0.62$ $[-0.85, -0.39]$ points below Exact Narrow ($90.52\%$ vs.\ $91.14\%$), whereas an intermediate allocation ($M{=}5{,}000, q{=}738$) gains $+0.78$ $[+0.48, +1.08]$ points ($91.92\%$). Even against a stronger \textbf{Packed Exact Narrow} baseline that stores only the upper triangle $\frac{1}{2}M_n(M_n{+}1)$ (widening $M_n$ by $\sqrt{2}\approx 1.41\times$), \method maintains a clear lead of $+1.06\text{ to }+1.65$ points on CUB-200 and VTAB-50 (Appendix~\ref{sec:app-packed}).
\end{itemize}

\begin{table}[t]
    \centering
    \footnotesize
    \setlength{\tabcolsep}{1.6pt}
    \caption{Iso-memory comparison at $\Bcal{=}36$ MiB under hold-out validation and refit ($3$ seeds, $\bar{\Acal}$ in \%). Unrestored LR, FD, and RFD receive $q{+}1$ rows to match the $M$ variance scalars of \method. \methodcal applies rank-calibrated hold-out selection.}
    \label{tab:budget}
    \begin{tabular*}{\linewidth}{@{\extracolsep{\fill}}lcccccc@{}}
        \toprule
        & \multicolumn{2}{c}{CIFAR-100} & \multicolumn{2}{c}{CUB-200} & \multicolumn{2}{c}{VTAB-50} \\
        \cmidrule(lr){2-3}\cmidrule(lr){4-5}\cmidrule(lr){6-7}
        Readout Rule & $\bar{\Acal}$ & $\Delta_{\mathrm{nar}}$ & $\bar{\Acal}$ & $\Delta_{\mathrm{nar}}$ & $\bar{\Acal}$ & $\Delta_{\mathrm{nar}}$ \\
        \midrule
        Exact Narrow ($M_n$) & 91.62 & 0.00 & 91.44 & 0.00 & 93.81 & 0.00 \\
        \midrule
        Unrestored LR & 91.69 & \up{0.07} & 86.12 & \down{5.32} & 94.21 & \up{0.40} \\
        Frequent Directions & 91.81 & \up{0.19} & 86.34 & \down{5.10} & 94.02 & \up{0.21} \\
        Robust FD (RFD) & \textbf{92.22} & \upb{0.60} & 91.88 & \upb{0.44} & \textbf{94.56} & \upb{0.75} \\
        FD + Exact Var. & 92.19 & \upb{0.57} & 91.58 & \up{0.14} & 94.35 & \up{0.54} \\
        \method (std.\ val.) & 92.20 & \upb{0.58} & 88.68 & \down{2.76} & 94.40 & \up{0.59} \\
        \textbf{\methodcal (ours)} & \underline{92.21} & \upb{0.59} & \textbf{91.89} & \upb{0.45} & \underline{94.41} & \upb{0.60} \\
        \midrule
        Unbudgeted Wide & 92.35 & +0.73 & 92.02 & +0.58 & 94.88 & +1.07 \\
        \bottomrule
    \end{tabular*}
\end{table}

Table~\ref{tab:budget} evaluates the same $\Bcal{=}36$ MiB envelope when $\gamma_t \in \{10^2,\ldots,10^9\}$ is chosen by hold-out validation and refit. On CIFAR-100, variance-corrected rules (RFD, FD+var, \method, \methodcal) all achieve significant gains of $+0.57\text{ to }+0.60$ points over Exact Narrow, recovering $>80\%$ of the unbudgeted wide ceiling ($92.35\%$), whereas unrestored LR ($+0.07$) does not. On CUB-200, however, standard hold-out validation triggers the Theorem~\ref{thm:amp} pathology: because Session-1 fit splits ($n_{\mathrm{fit}} \approx 850$) barely fit inside $q$ while refits ($n_{\mathrm{refit}} \approx 945$) exceed $q$, standard validation picks $\gamma_1{=}10^2$ on two seeds and collapses Unrestored LR ($86.12\%$), FD ($86.34\%$), and uncalibrated \method ($88.68\%$). RFD avoids collapse via its isotropic shrinkage floor $\alpha$, and \textbf{\methodcal} eliminates the collapse at the source by calibrating the validation truncation rank ($q_{\mathrm{val}}$), restoring CUB-200 accuracy to $\mathbf{91.89\%}$ ($+0.45$ over Exact Narrow).

\subsection{Mechanism and Component Ablations}
\label{sec:mechanism}

\begin{table}[t]
    \centering
    \footnotesize
    \setlength{\tabcolsep}{3.0pt}
    \caption{Ablation of retained sufficient-statistic components ($M{=}10{,}000$, $q{=}840$, shared GCV grid, $5$ seeds). Reporting both $\bar{\Acal}$ and $\Acal_T$ (\%) exposes the intermediate-vs.-final trade-off of span restriction.}
    \label{tab:ablation}
    \begin{tabular*}{\linewidth}{@{\extracolsep{\fill}}lcccccc@{}}
        \toprule
        & \multicolumn{2}{c}{CIFAR-100} & \multicolumn{2}{c}{CUB-200} & \multicolumn{2}{c}{VTAB-50} \\
        \cmidrule(lr){2-3}\cmidrule(lr){4-5}\cmidrule(lr){6-7}
        State Configuration & $\bar{\Acal}$ & $\Acal_T$ & $\bar{\Acal}$ & $\Acal_T$ & $\bar{\Acal}$ & $\Acal_T$ \\
        \midrule
        Diagonal $(v_t, \Gamma_t)$ only & 83.64 & 77.89 & 63.62 & 52.94 & 36.82 & 31.78 \\
        Subspace PCR ($\Pi_t W_t$) & 92.19 & 87.76 & 91.42 & 86.48 & 89.98 & \textbf{92.82} \\
        Unrestored LR ($F_t, \Gamma_t$) & 92.18 & 87.09 & 91.50 & 87.34 & 95.04 & 92.48 \\
        Frequent Directions (FD) & 91.80 & 85.82 & \textbf{91.64} & \textbf{87.44} & \textbf{95.08} & 92.61 \\
        Robust FD (RFD) & 92.65 & 88.49 & 91.31 & 86.72 & 95.06 & 92.55 \\
        \textbf{\method ($F_t, v_t, \Gamma_t$)} & \textbf{92.81} & \textbf{88.78} & 91.52 & 87.26 & 95.07 & 92.52 \\
        \bottomrule
    \end{tabular*}
\end{table}

\paragraph{What must the state retain?}
Table~\ref{tab:ablation} ablates the components of $(F_t, v_t, \Gamma_t)$ under a shared GCV grid. Discarding off-diagonal interactions entirely (\emph{Diagonal only}, $\Sighat_t = \Diag(v_t)$) degrades $\bar{\Acal}$ by $9.17$, $27.90$, and $58.25$ points across the three benchmarks, confirming that cross-coordinate co-activations are indispensable under standard regularization scales. Restricting the readout to the retained subspace (\emph{Subspace PCR}, which projects $\Gamma_t \leftarrow \Pi_t \Gamma_t$) drops VTAB-50 stage-average accuracy by $5.09$ points ($89.98\%$ vs.\ $95.07\%$) while slightly raising final accuracy $\Acal_T$ ($92.82\%$ vs.\ $92.52\%$), demonstrating that off-subspace cross-correlations $\Gamma_\perp$ are vital during early sessions before the rank-$q$ factor spans the full class geometry. Restoring exact coordinate variances in \method improves CIFAR-100 by $+0.63$ points in $\bar{\Acal}$ and $+1.69$ points in $\Acal_T$ over Unrestored LR on the exact same factor $F_t$.

\begin{figure}[t]
    \centering
    \includegraphics[width=\linewidth]{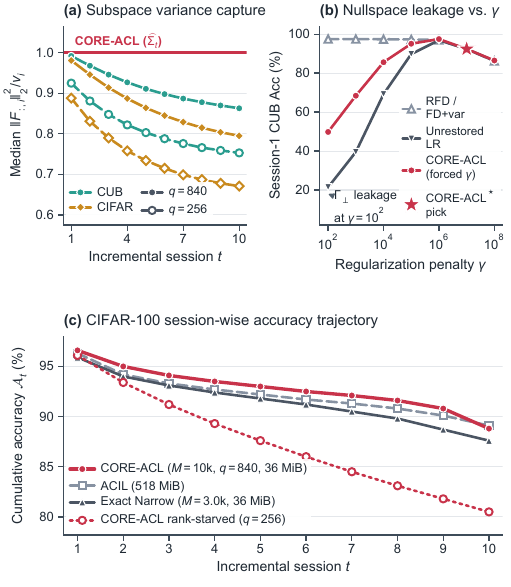}
    \caption{Mechanism diagnostics. \textbf{(a)} Median fraction of coordinate self-variance $\|F_{:,i}\|_2^2 / v_i$ captured by the rank-$q$ factor across sessions; \method restores the diagonal to $1.0$ exactly. \textbf{(b)} Session-1 CUB-200 accuracy across forced penalties $\gamma$: small $\gamma$ triggers $\Gamma_\perp$ blow-up in unrestored LR and uncalibrated \method, whereas \methodcal selects $\gamma{=}10^7$ (star). \textbf{(c)} Cumulative accuracy trajectory $\Acal_t$ across $10$ CIFAR-100 sessions.}
    \label{fig:mechanism}
\end{figure}

\paragraph{Variance erosion and nullspace amplification.}
Figure~\ref{fig:mechanism}a tracks the median variance capture ratio $\|F_{t,:,i}\|_2^2 / v_{t,i}$ across sessions $t{=}1,\ldots,10$. As new classes rotate the feature distribution, a fixed rank $q$ retains a steadily shrinking share of coordinate variance, falling to $0.671$ on CIFAR-100 at $q{=}256$ ($0.795$ at $q{=}840$). Without variance restoration, this lost energy directly weakens diagonal conditioning and causes the $8.18$-point late-session accuracy deficit seen in Figure~\ref{fig:results}a. Meanwhile, Figure~\ref{fig:mechanism}b visualizes Theorem~\ref{thm:amp} on Session $1$ of CUB-200: although off-subspace label energy accounts for $<0.035\%$ of $\|\Gamma_1\|_F$, forcing $\gamma{=}10^2$ amplifies $\Gamma_\perp$ until $\|(\Sighat_1 + \gamma I)^{-1}\Gamma_\perp\|_F / \|W_1\|_F > 0.998$, collapsing accuracy to $21.4\%$ (LR) and $49.8\%$ (\method). \methodcal detects this amplification on the pre-truncated validation split and selects $\gamma{=}10^7$ ($92.4\%$).

\subsection{Generalization and Computational Efficiency}
\label{sec:cost}

\begin{figure}[t]
    \centering
    \includegraphics[width=\linewidth]{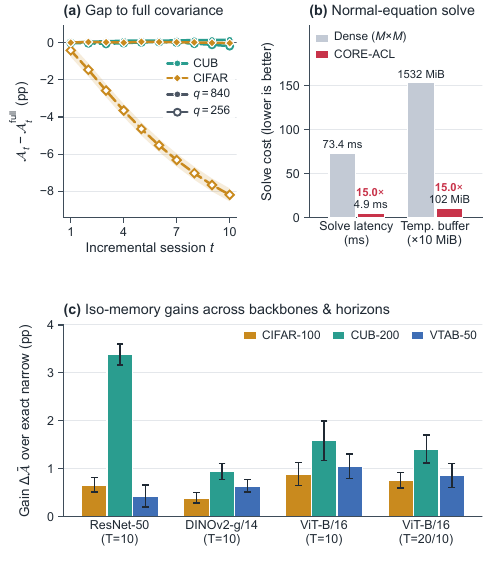}
    \caption{Stream fidelity, solver speedup, and cross-backbone transfer. \textbf{(a)} Session-wise gap $\Acal_t - \Acal_t^{\mathrm{full}}$ relative to dense full covariance. \textbf{(b)} Normal-equation solve latency and temporary buffer on RTX 3090 ($15.0\times$ speedup and memory reduction). \textbf{(c)} Paired $\Delta\bar{\Acal}$ gains of \method over state-matched Exact Narrow across ResNet-50, DINOv2-g/14, and $20$-session horizons.}
    \label{fig:results}
\end{figure}

\paragraph{Backbone and horizon generality.}
Figure~\ref{fig:results}c and Appendix Table~\ref{tab:app-encoder} verify that the width-over-rank allocation holds across diverse visual backbones and stream lengths at matched $36$ MiB state. On ResNet-50 \cite{he2016resnet}, where raw features are less linearly separable, \method beats state-matched Exact Narrow by $+0.66$, $+3.38$, and $+0.43$ points on CIFAR-100, CUB-200, and VTAB-50. On DINOv2 ViT-g/14 ($D{=}1536$) \cite{oquab2024dinov2}, gains remain consistently positive ($+0.39$, $+0.94$, $+0.64$), and on longer $20$- and $50$-session streams \method matches dense Fly-CL within $0.06$ points ($92.76\%$ vs.\ $92.80\%$ on $50$-session CIFAR-100).

\paragraph{Compute vs.\ memory trade-off.}
As shown in Figure~\ref{fig:results}b, the Woodbury normal-equation solve in Eq.~\eqref{eq:woodbury} takes $4.9$ ms and $102$ MiB of temporary buffer versus $73.4$ ms and $1{,}532$ MiB for dense $M\times M$ Cholesky factorization---a $15.0\times$ reduction in both latency and solver workspace. End-to-end session fitting on CIFAR-100 takes $11.4$ s per session for \method (dominated by the session SVD $\Trunc(\Phi_t)$) versus $24.6$ s for RFD and $0.8$ s for Exact Narrow (Appendix~\ref{sec:app-cost}). Thus, \method specifically optimizes persistent inter-session memory and normal-equation solve cost rather than feature-extraction or SVD wall-clock time.

\section{Conclusion}
\label{sec:conclusion}

We investigated how a fixed persistent memory budget in analytic continual learning should be allocated between feature expansion width and covariance interaction fidelity. Because coordinate self-variances and class cross-correlations scale linearly with width whereas off-diagonal co-activations scale quadratically, \method preserves the linear statistics exactly, compresses interactions into a streaming rank-$q$ factor, and restores exact diagonal variances via a fast Woodbury readout. Across five CIL benchmarks, \method matches dense $385$ MiB covariance readouts within $0.09$ points in just $35.9$ MiB, and demonstrates that spending bytes on width consistently beats narrow exactness once rank starvation and nullspace refit amplification (\methodcal) are controlled.

{\small
\bibliographystyle{ieeenat_fullname}
\bibliography{references}

@article{zhou2024continual,
  title={Continual Learning with Pre-Trained Models: A Survey},
  author={Zhou, Da-Wei and Sun, Hai-Long and Ning, Jingyi and Ye, Han-Jia and Zhan, De-Chuan},
  journal={arXiv preprint arXiv:2401.16386},
  year={2024}
}

@article{hou2026transition,
  title={Continual Learning in Transition},
  author={Hou, Zhiyan and Zhang, Dan and Feng, Tao and Wang, Liyuan and Li, Wei and Hao, Xiangzhao and An, Hongyan and Fang, Junfeng and Ma, Haokai and Xu, Zhaohui and Tang, Xinyu and Guo, Haiyun and Wang, Jinqiao and Chua, Tat-Seng},
  journal={arXiv preprint arXiv:2608.06216},
  year={2026}
}

@inproceedings{zhuang2022acil,
  title={{ACIL}: Analytic Class-Incremental Learning with Absolute Memorization and Privacy Protection},
  author={Zhuang, Huiping and Weng, Zhenyu and Wei, Hongxin and Xie, Renchunzi and Toh, Kar-Ann and Lin, Zhiping},
  booktitle={Advances in Neural Information Processing Systems (NeurIPS)},
  volume={35},
  pages={11602--11615},
  year={2022}
}

@inproceedings{mcdonnell2023ranpac,
  title={{RanPAC}: Random Projections and Pre-trained Models for Continual Learning},
  author={McDonnell, Mark D. and Gong, Dong and Parvaneh, Amin and Abbasnejad, Ehsan and van den Hengel, Anton},
  booktitle={Advances in Neural Information Processing Systems (NeurIPS)},
  volume={36},
  year={2023}
}

@inproceedings{zhuang2024dsal,
  title={{DS-AL}: A Dual-Stream Analytic Learning for Exemplar-Free Class-Incremental Learning},
  author={Zhuang, Huiping and He, Run and Tong, Kai and Zeng, Ziqian and Chen, Cen and Lin, Zhiping},
  booktitle={Proceedings of the AAAI Conference on Artificial Intelligence},
  volume={38},
  number={15},
  pages={17237--17244},
  year={2024},
  doi={10.1609/aaai.v38i15.29670}
}

@inproceedings{zhuang2024gacl,
  title={{GACL}: Exemplar-Free Generalized Analytic Continual Learning},
  author={Zhuang, Huiping and Chen, Yizhu and Fang, Di and He, Run and Tong, Kai and Wei, Hongxin and Zeng, Ziqian and Chen, Cen},
  booktitle={Advances in Neural Information Processing Systems (NeurIPS)},
  volume={37},
  year={2024}
}

@inproceedings{zhuang2024foal,
  title={{F-OAL}: Forward-only Online Analytic Learning with Fast Training and Low Memory Footprint in Class Incremental Learning},
  author={Zhuang, Huiping and Liu, Yuchen and He, Run and Tong, Kai and Zeng, Ziqian and Chen, Cen and Wang, Yi and Chau, Lap-Pui},
  booktitle={Advances in Neural Information Processing Systems (NeurIPS)},
  volume={37},
  pages={41517--41538},
  year={2024}
}

@article{he2024real,
  title={{REAL}: Representation Enhanced Analytic Learning for Exemplar-Free Class-Incremental Learning},
  author={He, Run and Fang, Di and Chen, Yizhu and Tong, Kai and Chen, Cen and Wang, Yi and Chau, Lap-Pui and Zhuang, Huiping},
  journal={arXiv preprint arXiv:2403.13522},
  year={2024}
}

@inproceedings{zou2026flycl,
  title={{Fly-CL}: A Fly-Inspired Framework for Enhancing Efficient Decorrelation and Reduced Training Time in Pre-trained Model-based Continual Representation Learning},
  author={Zou, Heming and Zang, Yunliang and Xu, Wutong and Ji, Xiangyang},
  booktitle={International Conference on Learning Representations (ICLR)},
  year={2026}
}

@inproceedings{tran2025mvp,
  title={Boosting Multiple Views for Pretrained-Based Continual Learning},
  author={Tran, Quyen and Tran, Lam and Doan, Khanh and Tran, Toan and Phung, Dinh and Than, Khoat and Le, Trung},
  booktitle={International Conference on Learning Representations (ICLR)},
  year={2025}
}

@inproceedings{li2026grp,
  title={Enhancing Pretrained Model-Based Continual Representation Learning via Guided Random Projection},
  author={Li, Ruilin and Zou, Heming and Yan, Xiufeng and Liang, Zheming and Yang, Jie and Li, Chenliang and Yang, Xue},
  booktitle={European Conference on Computer Vision (ECCV)},
  pages={246--264},
  year={2026}
}

@inproceedings{peng2025loranpac,
  title={{LoRanPAC}: Low-rank Random Features and Pre-trained Models for Bridging Theory and Practice in Continual Learning},
  author={Peng, Liangzu and Elenter, Juan and Agterberg, Joshua and Ribeiro, Alejandro and Vidal, Ren{\'e}},
  booktitle={International Conference on Learning Representations (ICLR)},
  year={2025}
}

@article{ghashami2016fd,
  title={Frequent Directions: Simple and Deterministic Matrix Sketching},
  author={Ghashami, Mina and Liberty, Edo and Phillips, Jeff M. and Woodruff, David P.},
  journal={SIAM Journal on Computing},
  volume={45},
  number={5},
  pages={1762--1792},
  year={2016},
  doi={10.1137/15M1009718}
}

@article{luo2019rfd,
  title={Robust Frequent Directions with Application in Online Learning},
  author={Luo, Luo and Chen, Cheng and Zhang, Zhihua and Li, Wu-Jun and Zhang, Tong},
  journal={Journal of Machine Learning Research},
  volume={20},
  number={45},
  pages={1--41},
  year={2019}
}

@inproceedings{shi2021sketch,
  title={A Deterministic Streaming Sketch for Ridge Regression},
  author={Shi, Benwei and Phillips, Jeff M.},
  booktitle={International Conference on Artificial Intelligence and Statistics (AISTATS)},
  pages={586--594},
  year={2021}
}

@article{golub1979gcv,
  title={Generalized Cross-Validation as a Method for Choosing a Good Ridge Parameter},
  author={Golub, Gene H. and Heath, Michael and Wahba, Grace},
  journal={Technometrics},
  volume={21},
  number={2},
  pages={215--223},
  year={1979}
}

@inproceedings{wang2022l2p,
  title={Learning to Prompt for Continual Learning},
  author={Wang, Zifeng and Zhang, Zizhao and Lee, Chen-Yu and Zhang, Han and Sun, Ruoxi and Ren, Xiaoqi and Su, Guolong and Perot, Vincent and Dy, Jennifer and Pfister, Tomas},
  booktitle={Proceedings of the IEEE/CVF Conference on Computer Vision and Pattern Recognition (CVPR)},
  pages={139--149},
  year={2022}
}

@inproceedings{wang2022dualprompt,
  title={{DualPrompt}: Complementary Prompting for Rehearsal-Free Continual Learning},
  author={Wang, Zifeng and Zhang, Zizhao and Ebrahimi, Sayna and Sun, Ruoxi and Zhang, Han and Lee, Chen-Yu and Ren, Xiaoqi and Su, Guolong and Perot, Vincent and Dy, Jennifer and Pfister, Tomas},
  booktitle={European Conference on Computer Vision (ECCV)},
  pages={631--648},
  year={2022}
}

@inproceedings{smith2023codaprompt,
  title={{CODA-Prompt}: {CO}ntinual Decomposed Attention-Based Prompting for Rehearsal-Free Continual Learning},
  author={Smith, James Seale and Karlinsky, Leonid and Gutta, Vyshnavi and Cascante-Bonilla, Paola and Kim, Donghyun and Arbelle, Assaf and Panda, Rameswar and Feris, Rogerio and Kira, Zsolt},
  booktitle={Proceedings of the IEEE/CVF Conference on Computer Vision and Pattern Recognition (CVPR)},
  pages={11909--11919},
  year={2023}
}

@inproceedings{zhou2024ease,
  title={Expandable Subspace Ensemble for Pre-Trained Model-Based Class-Incremental Learning},
  author={Zhou, Da-Wei and Sun, Hai-Long and Ye, Han-Jia and Zhan, De-Chuan},
  booktitle={Proceedings of the IEEE/CVF Conference on Computer Vision and Pattern Recognition (CVPR)},
  pages={23554--23564},
  year={2024}
}

@inproceedings{liang2024inflora,
  title={{InfLoRA}: Interference-Free Low-Rank Adaptation for Continual Learning},
  author={Liang, Yan-Shuo and Li, Wu-Jun},
  booktitle={Proceedings of the IEEE/CVF Conference on Computer Vision and Pattern Recognition (CVPR)},
  pages={23638--23647},
  year={2024}
}

@inproceedings{sun2025mos,
  title={{MOS}: Model Surgery for Pre-Trained Model-Based Class-Incremental Learning},
  author={Sun, Hai-Long and Zhou, Da-Wei and Zhao, Hanbin and Gan, Le and Zhan, De-Chuan and Ye, Han-Jia},
  booktitle={Proceedings of the AAAI Conference on Artificial Intelligence},
  volume={39},
  number={19},
  pages={20699--20707},
  year={2025}
}

@inproceedings{wang2025self,
  title={Self-Expansion of Pre-Trained Models with Mixture of Adapters for Continual Learning},
  author={Wang, Huiyi and Lu, Haodong and Yao, Lina and Gong, Dong},
  booktitle={Proceedings of the IEEE/CVF Conference on Computer Vision and Pattern Recognition (CVPR)},
  pages={10087--10098},
  year={2025}
}

@inproceedings{zou2025flylora,
  title={{FlyLoRA}: Boosting Task Decoupling and Parameter Efficiency via Implicit Rank-Wise Mixture-of-Experts},
  author={Zou, Heming and Zang, Yunliang and Xu, Wutong and Zhu, Yao and Ji, Xiangyang},
  booktitle={Advances in Neural Information Processing Systems (NeurIPS)},
  volume={38},
  pages={10386--10419},
  year={2025}
}

@article{wang2025harmonious,
  title={Harmonious Parameter Adaptation in Continual Visual Instruction Tuning for Safety-Aligned {MLLMs}},
  author={Wang, Ziqi and Che, Chang and Wang, Qi and Ma, Hui and Shi, Zenglin and Snoek, Cees G. M. and Wang, Meng},
  journal={arXiv preprint arXiv:2511.20158},
  year={2025}
}

@article{jiang2026miles,
  title={Miles: Metric Learning with Expandable Subspace for Pre-Trained Model-Based Class-Incremental Learning},
  author={Jiang, Kai and Lin, Zisong and Zhang, Hongyuan and Bai, Xueru and Li, Xuelong},
  journal={arXiv preprint arXiv:2607.17593},
  year={2026}
}

@article{xu2026geometry,
  title={Geometry-Anchored Transport Framework for Exemplar-Free Class-Incremental Learning},
  author={Xu, Hongye and Krawczyk, Bartosz},
  journal={arXiv preprint arXiv:2606.25347},
  year={2026}
}

@article{hu2026visual,
  title={Visual Branch is What You Need for {CLIP}-based Class-Incremental Learning},
  author={Hu, Tao and Xie, Zhen-Hao and Guo, Jingcai and Zhan, De-Chuan and Zhou, Da-Wei},
  journal={arXiv preprint arXiv:2609.37888},
  year={2026}
}

@inproceedings{prabhu2024randumb,
  title={{RanDumb}: Random Representations Outperform Online Continually Learned Representations},
  author={Prabhu, Ameya and Sinha, Shiven and Kumaraguru, Ponnurangam and Torr, Philip H. S. and Sener, Ozan and Dokania, Puneet K.},
  booktitle={Advances in Neural Information Processing Systems (NeurIPS)},
  volume={37},
  year={2024}
}

@inproceedings{cao2026statistics,
  title={On the Power of Statistics in Class-Incremental Learning with Pretrained Models},
  author={Cao, Zhiwen and Li, Yanfeng and Huang, Shu-Dong and Ye, Yalan and Xia, Shuyin and Wang, Yi and Lv, Jiancheng},
  booktitle={Proceedings of the 43rd International Conference on Machine Learning (ICML)},
  volume={306},
  pages={11255--11271},
  year={2026}
}

@inproceedings{momeni2025anacp,
  title={{AnaCP}: Toward Upper-Bound Continual Learning via Analytic Contrastive Projection},
  author={Momeni, Saleh and Xiao, Changnan and Liu, Bing},
  booktitle={Advances in Neural Information Processing Systems (NeurIPS)},
  volume={38},
  year={2025}
}

@article{zou2025structural,
  title={Structural Features of the Fly Olfactory Circuit Mitigate the Stability-Plasticity Dilemma in Continual Learning},
  author={Zou, Heming and Zang, Yunliang and Ji, Xiangyang},
  journal={arXiv preprint arXiv:2502.01427},
  year={2025}
}

@article{tran2026spectral,
  title={Spectral-Aware Analytic Class-Incremental Learning for Long-Tailed Distributions},
  author={Tran, Quyen and Nguyen, Hai and Dao, Quan and Li, Zhuowei and Le, Nam and Le, Trung and Metaxas, Dimitris},
  journal={arXiv preprint arXiv:2607.22931},
  year={2026}
}

@article{sezer2026impossibility,
  title={An Impossibility Result for {ETF} Target Geometry, and an Equinorm Remedy for Imbalanced Analytic Continual Learning},
  author={Sezer, Berk},
  journal={arXiv preprint arXiv:2610.24844},
  year={2026}
}

@article{mao2026trace,
  title={{TRACE}: Memory-Efficient Time Series Class-Incremental Learning via Temporal-Spectral Reconstruction},
  author={Mao, Junhui and Zhang, Junru and Ye, Liming and Peng, Cheng and Shi, Haoran and Tang, Xiaoli and Wang, Baokun and Dou, Bin and Dong, Yabo and Yu, Han},
  journal={arXiv preprint arXiv:2610.35727},
  year={2026}
}

@article{zhuang2024drive,
  title={Online Analytic Exemplar-Free Continual Learning with Large Models for Imbalanced Autonomous Driving Task},
  author={Zhuang, Huiping and Fang, Di and Tong, Kai and Liu, Yuchen and Zeng, Ziqian and Zhou, Xu and Chen, Cen},
  journal={arXiv preprint arXiv:2405.17779},
  year={2024}
}

@article{song2026drift,
  title={Analytic Drift Resister for Non-Exemplar Continual Graph Learning},
  author={Song, Lei and Guan, Shihan and Kong, Youyong},
  journal={arXiv preprint arXiv:2604.02633},
  year={2026}
}

@inproceedings{xiao2025analytickws,
  title={{AnalyticKWS}: Towards Exemplar-Free Analytic Class Incremental Learning for Small-footprint Keyword Spotting},
  author={Xiao, Yang and Peng, Tianyi and Das, Rohan Kumar and Hu, Yuchen and Zhuang, Huiping},
  booktitle={Findings of the Association for Computational Linguistics: ACL 2025},
  pages={14147--14158},
  year={2025},
  doi={10.18653/v1/2025.findings-acl.728}
}

@inproceedings{snelson2005spgp,
  title={Sparse Gaussian Processes Using Pseudo-Inputs},
  author={Snelson, Edward and Ghahramani, Zoubin},
  booktitle={Advances in Neural Information Processing Systems (NeurIPS)},
  volume={18},
  pages={1257--1264},
  year={2005}
}

@article{krizhevsky2009cifar,
  title={Learning Multiple Layers of Features from Tiny Images},
  author={Krizhevsky, Alex},
  journal={Technical Report, University of Toronto},
  year={2009}
}

@article{wah2011cub,
  title={The {Caltech-UCSD Birds-200-2011} Dataset},
  author={Wah, Catherine and Branson, Steve and Welinder, Peter and Perona, Pietro and Belongie, Serge},
  journal={Technical Report CNS-TR-2011-001, California Institute of Technology},
  year={2011}
}

@article{zhai2019vtab,
  title={A Large-Scale Study of Representation Learning with the Visual Task Adaptation Benchmark},
  author={Zhai, Xiaohua and Puigcerver, Joan and Kolesnikov, Alexander and Ruyssen, Pierre and Riquelme, Carlos and Lucic, Mario and Djolonga, Josip and Pinto, Andre Susano and Neumann, Maxim and Dosovitskiy, Alexey and Beyer, Lucas and Bachem, Olivier and Tschannen, Michael and Michalski, Marcin and Bousquet, Olivier and Gelly, Sylvain and Houlsby, Neil},
  journal={arXiv preprint arXiv:1910.04867},
  year={2019}
}

@inproceedings{hendrycks2021imagenetr,
  title={The Many Faces of Robustness: A Critical Analysis of Out-of-Distribution Generalization},
  author={Hendrycks, Dan and Basart, Steven and Mu, Norman and Kadavath, Saurav and Wang, Frank and Dorundo, Evan and Desai, Rahul and Zhu, Tyler and Parajuli, Samyak and Guo, Mike and Song, Dawn and Steinhardt, Jacob and Gilmer, Justin},
  booktitle={Proceedings of the IEEE/CVF International Conference on Computer Vision (ICCV)},
  pages={8340--8349},
  year={2021}
}

@inproceedings{hendrycks2021imageneta,
  title={Natural Adversarial Examples},
  author={Hendrycks, Dan and Zhao, Kevin and Basart, Steven and Steinhardt, Jacob and Song, Dawn},
  booktitle={Proceedings of the IEEE/CVF Conference on Computer Vision and Pattern Recognition (CVPR)},
  pages={15262--15271},
  year={2021}
}

@inproceedings{dosovitskiy2021vit,
  title={An Image is Worth 16x16 Words: Transformers for Image Recognition at Scale},
  author={Dosovitskiy, Alexey and Beyer, Lucas and Kolesnikov, Alexander and Weissenborn, Dirk and Zhai, Xiaohua and Unterthiner, Thomas and Dehghani, Mostafa and Minderer, Matthias and Heigold, Georg and Gelly, Sylvain and Uszkoreit, Jakob and Houlsby, Neil},
  booktitle={International Conference on Learning Representations (ICLR)},
  year={2021}
}

@inproceedings{he2016resnet,
  title={Deep Residual Learning for Image Recognition},
  author={He, Kaiming and Zhang, Xiangyu and Ren, Shaoqing and Sun, Jian},
  booktitle={Proceedings of the IEEE Conference on Computer Vision and Pattern Recognition (CVPR)},
  pages={770--778},
  year={2016}
}

@article{oquab2024dinov2,
  title={{DINOv2}: Learning Robust Visual Features without Supervision},
  author={Oquab, Maxime and Darcet, Timoth{\'e}e and Moutakanni, Th{\'e}o and Vo, Huy V. and Szafraniec, Marc and Khalidov, Vasil and Fernandez, Pierre and Haziza, Daniel and Massa, Francisco and El-Nouby, Alaaeldin and Assran, Mahmoud and Ballas, Nicolas and Galuba, Wojciech and Howes, Russell and Huang, Po-Yao and Li, Shang-Wen and Misra, Ishan and Rabbat, Michael and Sharma, Vasu and Synnaeve, Gabriel and Xu, Hu and J{\'e}gou, Herv{\'e} and Mairal, Julien and Labatut, Patrick and Joulin, Armand and Bojanowski, Piotr},
  journal={Transactions on Machine Learning Research},
  year={2024}
}
}

\clearpage
\appendix
\onecolumn
\section{Mathematical Proofs and Extended Theory}
\label{sec:app-proofs}

Throughout this appendix, we assume exact arithmetic, a fixed feature lifting map $\phi_M:\mathcal{X}\to\R^M$, a class vocabulary of size $C\ge 2$, and ridge regularization parameter $\gamma>0$. Rank-deficient session matrices and empty initial factors are permitted. For any matrix $X\in\R^{\ell\times M}$, let $b(X)=\min(q,\mathrm{rank}(X),M)$ denote the number of singular rows retained by $\Trunc(X)$, and write $q_t = b(M_t)$ where $M_t = [F_{t-1}; U_t]$ is the stacked factor at session $t$. The session truncation residual and the merge truncation residual are defined as
\begin{equation}
\Omega_t^{\mathrm{sess}} = \Phi_t^\top\Phi_t - U_t^\top U_t,
\qquad
\Omega_t^{\mathrm{merge}} = M_t^\top M_t - F_t^\top F_t,
\qquad
M_t = \begin{bmatrix} F_{t-1}\\ U_t \end{bmatrix}.
\end{equation}
Both $\Omega_t^{\mathrm{sess}}$ and $\Omega_t^{\mathrm{merge}}$ are positive semidefinite (PSD) by the Eckart--Young--Mirsky theorem, since each equals a sum of outer products of discarded right singular vectors weighted by squared singular values.

\subsection{Proof of Proposition~\ref{prop:psd} (Loewner Positivity, Operator Sandwich, and Optimal Diagonal Restoration)}

\begin{proof}
We establish $F_t^\top F_t \preceq \Sigma_t$ by induction over sessions $t\ge 0$. At $t=0$, $F_0 = 0$ and $\Sigma_0 = 0$. Suppose $F_{t-1}^\top F_{t-1} \preceq \Sigma_{t-1}$. Because $U_t = \Trunc(\Phi_t)$ discards only non-negative singular modes, $U_t^\top U_t \preceq \Phi_t^\top \Phi_t$. By block-matrix multiplication,
\begin{equation}
M_t^\top M_t = F_{t-1}^\top F_{t-1} + U_t^\top U_t \preceq \Sigma_{t-1} + \Phi_t^\top \Phi_t = \Sigma_t.
\end{equation}
Applying $\Trunc$ to $M_t$ again drops a PSD tail $\Omega_t^{\mathrm{merge}}\succeq 0$, yielding $F_t^\top F_t = M_t^\top M_t - \Omega_t^{\mathrm{merge}} \preceq M_t^\top M_t \preceq \Sigma_t$. Consequently, the cumulative discarded covariance satisfies
\begin{equation}
\Omega_t = \Sigma_t - F_t^\top F_t = \Omega_{t-1} + \Omega_t^{\mathrm{sess}} + \Omega_t^{\mathrm{merge}} = \sum_{j=1}^{t}\big(\Omega_j^{\mathrm{sess}} + \Omega_j^{\mathrm{merge}}\big) \succeq 0.
\end{equation}
Because every diagonal entry of a PSD matrix is non-negative, $\rho_{t,i} = (\Omega_t)_{i,i} \ge 0$ for all $i\in\{1,\ldots,M\}$. Thus $\Sighat_t = F_t^\top F_t + \Diag(\rho_t)$ is a sum of two PSD matrices and satisfies $\Sighat_t \succeq 0$. Since $v_t = \diag(\Sigma_t)$ is accumulated without approximation, $\diag(\Sighat_t) = \diag(F_t^\top F_t) + v_t - \diag(F_t^\top F_t) = v_t$.

For part (ii), write the approximation residual as $\Delta_t = \Sigma_t - \Sighat_t = \Omega_t - \Diag(\rho_t)$. Since $\Diag(\rho_t)\succeq 0$, we immediately have $\Delta_t \preceq \Omega_t$. Conversely, since $\Omega_t = \Delta_t + \Diag(\rho_t) \succeq 0$, we have $-\Diag(\rho_t) \preceq \Delta_t$. Because $0 \le \rho_{t,i} = e_i^\top \Omega_t e_i \le \|\Omega_t\|_2$, the extreme eigenvalues of $\Delta_t$ lie in $[-\|\Omega_t\|_2,\, \|\Omega_t\|_2]$, which proves $\|\Delta_t\|_2 \le \|\Omega_t\|_2 \le \trace(\Omega_t) = \sum_{i=1}^{M}\rho_{t,i}$.

For part (iii), fix $F_t$ and consider any candidate diagonal vector $u\in\R^M$. Decomposing $\Sigma_t - F_t^\top F_t - \Diag(u) = \Delta_t + \Diag(\rho_t - u)$, note that $\Delta_t$ has a strictly zero diagonal while $\Diag(\rho_t - u)$ is purely diagonal. Hence $\langle \Delta_t, \Diag(\rho_t - u)\rangle_F = 0$, and the Pythagorean identity in Frobenius norm gives
\begin{equation}
\norm{\Sigma_t - F_t^\top F_t - \Diag(u)}_F^2 = \norm{\Delta_t}_F^2 + \norm{\rho_t - u}_2^2,
\end{equation}
which is uniquely minimized at $u^\star = \rho_t$. Setting $u=0$ yields an excess Frobenius error of $\|\rho_t\|_2^2$, and restricting $u=c\mathbf{1}_M$ to an isotropic shift yields a minimum at $c=\bar{\rho}_t = \frac{1}{M}\sum_{i=1}^M \rho_{t,i}$ with excess Frobenius error $\|\rho_t - \bar{\rho}_t\mathbf{1}_M\|_2^2$.
\end{proof}

\subsection{Proof of Proposition~\ref{prop:reg} (Coordinate-Adaptive Regularization Equivalence)}

\begin{proof}
Expanding the Frobenius norm in Eq.~\eqref{eq:obj} trace-wise gives
\begin{equation}
\norm{F_t W}_F^2 + \sum_{i=1}^{M}(\gamma + \rho_{t,i})\norm{W_{i,:}}_2^2 - 2\langle \Gamma_t, W\rangle
=
\trace\!\Big(W^\top \big(F_t^\top F_t + \Diag(\rho_t) + \gamma I_M\big) W\Big) - 2\trace(W^\top \Gamma_t).
\end{equation}
Since $\Sighat_t + \gamma I_M = F_t^\top F_t + \Diag(\rho_t) + \gamma I_M \succ 0$ by Proposition~\ref{prop:psd}, the objective is strictly convex in $W$. Setting its matrix derivative $2(\Sighat_t + \gamma I_M)W - 2\Gamma_t = 0$ yields the unique global minimizer $W_t = (\Sighat_t + \gamma I_M)^{-1}\Gamma_t$. When $\rho_t$ is replaced by an isotropic scalar $\bar{\rho}_t\mathbf{1}_M$, the system matrix becomes $F_t^\top F_t + (\gamma + \bar{\rho}_t)I_M$, which coincides with the unrestored low-rank ridge operator $W_{\mathrm{LR}}(\gamma + \bar{\rho}_t)$.
\end{proof}

\subsection{Proof of Theorem~\ref{thm:amp} and Nullspace Stabilization Guarantees}

\begin{proof}
Let $F_t \in \R^{q_t\times M}$ have economy SVD $F_t = P_F S_F V_F^\top$ with positive singular values $\sigma_1 \ge \cdots \ge \sigma_r > 0$ ($r = \mathrm{rank}(F_t) \le q_t$). The orthogonal projector onto $\mathrm{row}(F_t) = \mathrm{col}(V_F)$ is $\Pi_t = V_F V_F^\top$. Because $F_t^\top F_t + \gamma I_M = V_F(S_F^2 + \gamma I_r)V_F^\top + \gamma(I_M - \Pi_t)$, its inverse acts as $V_F(S_F^2 + \gamma I_r)^{-1}V_F^\top$ on $\mathrm{col}(V_F)$ and as $\gamma^{-1}I$ on $\mathrm{col}(V_F)^\perp$. Applying this inverse to $\Gamma_t = \Gamma_\parallel + \Gamma_\perp$ gives Eq.~\eqref{eq:lr-split}, where the two terms lie in orthogonal subspaces column by column. A unit component along $v_{F,i}$ is scaled by $(\sigma_i^2 + \gamma)^{-1}$ whereas a unit component in $\mathrm{col}(V_F)^\perp$ is scaled by $\gamma^{-1}$, giving the relative ratio $(\sigma_i^2 + \gamma)/\gamma$. If the factor is exact ($F_t^\top F_t = \Sigma_t$), then $\mathrm{row}(F_t) = \mathrm{row}([\Phi_1;\ldots;\Phi_t])$, which contains the column space of $\Gamma_t = \sum_{j\le t}\Phi_j^\top Y_j$, so $\Gamma_\perp = 0$. Finally, for any unit vector $u \in \mathrm{row}(F_t)^\perp$, $F_t u = 0$, so $u^\top(\Sighat_t + \gamma I_M)u = u^\top\Diag(\rho_t)u + \gamma = \gamma + \sum_{i=1}^M \rho_{t,i} u_i^2$.
\end{proof}

\begin{proposition}[Why Spectral Shrinkage Floors Prevent Refit Blow-Up]
\label{prop:floor}
Consider a spectral sketch with an isotropic floor $\alpha > 0$, $\Sighat_t^{(\alpha)} = F_t^\top F_t + \alpha I_M$ (as in RFD \cite{luo2019rfd}). For any penalty $\gamma > 0$, off-subspace components in $\Gamma_\perp$ are scaled by $(\gamma + \alpha)^{-1}$, capping the relative amplification against mode $\sigma_i$ at
\begin{equation}
\mathcal{R}_{\mathrm{amp}}(\gamma;\alpha) = \frac{\sigma_i^2 + \gamma + \alpha}{\gamma + \alpha} \le 1 + \frac{\sigma_i^2}{\alpha}.
\end{equation}
In particular, even as $\gamma \to 0^+$, the amplification remains bounded by $1 + \sigma_i^2/\alpha < \infty$, whereas for unrestored low-rank ridge ($\alpha = 0$) it diverges as $\mathcal{O}(\gamma^{-1})$.
\end{proposition}

\begin{proposition}[Rank-Calibrated Validation Exposes Nullspace Sensitivity]
\label{prop:refit}
Suppose session $1$ has $n_{\mathrm{fit}} \le q < n_{\mathrm{refit}}$ and $\mathrm{rank}(\Phi_{\mathrm{fit}}) = n_{\mathrm{fit}}$. Standard validation at capacity $q$ retains all $n_{\mathrm{fit}}$ singular modes ($\Gamma_\perp^{\mathrm{val}} = 0$). Under \methodcal, the validation truncation rank $q_{\mathrm{val}} = \lfloor q \cdot n_{\mathrm{fit}} / n_{\mathrm{refit}} \rfloor$ satisfies $q_{\mathrm{val}} \le n_{\mathrm{fit}} - 1$, forcing at least $n_{\mathrm{fit}} - q_{\mathrm{val}} \ge 1$ non-zero singular directions into the validation nullspace so that $\Gamma_\perp^{\mathrm{val}} \neq 0$ whenever the held-out tail carries label correlation.
\end{proposition}

\begin{proposition}[Retrospective Margin Transfer Bound]
\label{prop:transfer}
Fit \method ($W_t$) and unbudgeted wide ridge ($W_t^\star$) on the same lifted features at the same session $t$ and penalty $\gamma > 0$. Then
\begin{equation}
W_t - W_t^\star = (\Sighat_t + \gamma I_M)^{-1}\Delta_t W_t^\star,
\qquad
\norm{W_t - W_t^\star}_F \le \frac{\norm{\Delta_t W_t^\star}_F}{\gamma + \min_i \rho_{t,i}} \le \frac{\trace(\Omega_t)\norm{\Gamma_t}_F}{\gamma(\gamma + \min_i \rho_{t,i})}.
\end{equation}
Let $\mu_{\mathrm{wide}}(z) = z^\top W_{t,:,y}^\star - \max_{c\neq y} z^\top W_{t,:,c}^\star$ denote the signed margin of $z = \phi_M(x)$ under $W_t^\star$, and define the margin-violation event $\Xi = \{|\mu_{\mathrm{wide}}(z)| \le 2\norm{z}_2\norm{W_t - W_t^\star}_F\}$. For any narrow exact readout $W_{\mathrm{narrow}}$ evaluated on the same test distribution,
\begin{equation}
\Acal_t(\method) - \Acal_t(\mathrm{narrow}) \ge \Acal_t(\mathrm{wide}) - \Acal_t(\mathrm{narrow}) - \Pr(\Xi).
\end{equation}
\end{proposition}

\begin{proof}
Using the resolvent identity $A^{-1} - B^{-1} = A^{-1}(B - A)B^{-1}$ with $A = \Sighat_t + \gamma I_M$ and $B = \Sigma_t + \gamma I_M$, we obtain $W_t - W_t^\star = (\Sighat_t + \gamma I_M)^{-1}(\Sigma_t - \Sighat_t)W_t^\star = (\Sighat_t + \gamma I_M)^{-1}\Delta_t W_t^\star$. Since $\Sighat_t + \gamma I_M \succeq (\gamma + \min_i \rho_{t,i})I_M$, submultiplicativity and Proposition~\ref{prop:psd}(ii) yield the norm bounds. For any test vector $z$, Cauchy--Schwarz gives $\max_c |z^\top(W_t - W_t^\star)_{:,c}| \le \|z\|_2\|W_t - W_t^\star\|_F$. Thus, outside $\Xi$, the top-scoring class of $W_t$ coincides with that of $W_t^\star$, establishing the accuracy transfer inequality.
\end{proof}

\section{Worked Two-Coordinate Analytical Example}
\label{sec:app-toy}

To see clearly why Theorem~\ref{thm:amp} distinguishes early-session refits from late-session updates, consider a $2$-dimensional lifted space ($M{=}2$) with rank-$1$ factor $F = (1,\, 0)$, ridge penalty $\gamma = 10^{-2}$, and class cross-correlation $\Gamma = (0,\, 2\times 10^{-2})^\top$, so that all label correlation lies orthogonal to $\mathrm{row}(F)$.
\begin{itemize}[leftmargin=1.3em]
    \item \textbf{Unrestored Low-Rank Ridge ($W_{\mathrm{LR}}$):} The inverted matrix is $\diag(1.01,\, 0.01)^{-1}$, yielding readout weights $W_{\mathrm{LR}} = (0,\, 2.0)^\top$. A tiny off-subspace cross-correlation of $0.02$ is amplified $100\times$ into a dominant weight of $2.0$.
    \item \textbf{\method in Late Stream ($\rho_2 = 0.50$, substantial discarded variance):} The variance-restored matrix is $\diag(1.01,\, 0.51)$, yielding $W_{\mathrm{CORE}} = (0,\, 0.02/0.51)^\top \approx (0,\, 0.039)^\top$. The restored self-variance acts as a natural coordinate stabilizer.
    \item \textbf{\method on Session-1 Refit ($\rho_2 = 0.01$, minimal discarded variance):} Because the factor $F_1$ just crossed the rank cap $q$, $99\%$ of coordinate variance is still inside $F_1$ ($\rho_2 = 0.01$). The system matrix is $\diag(1.01,\, 0.02)$, returning $W_{\mathrm{CORE}} = (0,\, 1.0)^\top$---still a $50\times$ blow-up. This explains why diagonal variance restoration alone cannot prevent Session-1 hold-out collapse at $\gamma = 10^2$, and why \methodcal's rank-calibrated validation split is essential.
\end{itemize}

\section{Algorithmic and Numerical Implementation Details}
\label{sec:app-alg}

\begin{algorithm}[h]
\small
\caption{Task-Local GCV Penalty Selection}
\label{alg:gcv}
\KwInput{Session features $\Phi_t \in \R^{n_t\times M}$, targets $Y_t \in \{0,1\}^{n_t\times C}$, economy SVD $\Phi_t = L_t S_t V_t^\top$, grid $\mathcal{G}$}
$\Psi_t \leftarrow L_t^\top Y_t$\;
\For{$\gamma \in \mathcal{G}$}{
  $\nu_i(\gamma) \leftarrow \sigma_i^2 / (\sigma_i^2 + \gamma)$;\quad
  $\mathrm{df}(\gamma) \leftarrow \sum_i \nu_i(\gamma)$\;
  $\widehat{Y}_t(\gamma) \leftarrow L_t \Diag(\nu(\gamma))\Psi_t$\;
  $\mathrm{GCV}(\gamma) \leftarrow \frac{\norm{Y_t - \widehat{Y}_t(\gamma)}_F^2 / n_t}{\big(1 - \mathrm{df}(\gamma)/n_t\big)^2}$\;
}
\Return $\arg\min_{\gamma\in\mathcal{G}} \mathrm{GCV}(\gamma)$ (breaking ties toward smaller $\gamma$)\;
\end{algorithm}

\begin{algorithm}[h]
\small
\caption{Hold-Out Validation with Rank Calibration (\methodcal)}
\label{alg:val}
\KwInput{Pre-session state $(F_{t-1}, v_{t-1}, \Gamma_{t-1})$, session data $(\Phi_t, Y_t)$, grid $\mathcal{G}$, target capacity $q$}
Partition $(\Phi_t, Y_t)$ into stratified $90\%$ fit split $(\Phi_{\mathrm{fit}}, Y_{\mathrm{fit}})$ ($n_{\mathrm{fit}}$ rows) and $10\%$ val split $(\Phi_{\mathrm{val}}, Y_{\mathrm{val}})$\;
\eIf{using \methodcal}{
  $q_{\mathrm{val}} \leftarrow \max\!\big(1, \lfloor q \cdot n_{\mathrm{fit}} / n_t \rfloor\big)$\tcp*{proportional truncation rank matching refit ratio}
}{
  $q_{\mathrm{val}} \leftarrow q$\tcp*{standard uncalibrated validation}
}
Form temporary state $(\tilde{F}_t, \tilde{v}_t, \tilde{\Gamma}_t)$ from $(F_{t-1}, v_{t-1}, \Gamma_{t-1})$ and $(\Phi_{\mathrm{fit}}, Y_{\mathrm{fit}})$ using truncation rank $q_{\mathrm{val}}$\;
Evaluate validation accuracy on $(\Phi_{\mathrm{val}}, Y_{\mathrm{val}})$ for every $\gamma \in \mathcal{G}$ via Eq.~\eqref{eq:woodbury}; select best $\gamma_t$\;
Discard temporary state; update $(F_t, v_t, \Gamma_t)$ on full $(\Phi_t, Y_t)$ at rank $q$ and solve for $W_t$ at $\gamma_t$\;
\end{algorithm}

\paragraph{Numerical precision and clipping safeguards.}
All persistent states $(F_t, v_t, \Gamma_t)$ are stored in IEEE-754 FP32. Before computing $\rho_t = [v_t - \diag(F_t^\top F_t)]_+$, we verify that $\min_i (v_{t,i} - \|F_{t,:,i}\|_2^2) \ge -256\,\varepsilon_{\mathrm{fp32}}\max(1, \|v_t\|_\infty)$ to ensure that any negative entry is purely floating-point rounding rather than a violation of Loewner positivity. During the Woodbury normal-equation solve (Eq.~\eqref{eq:woodbury}), $F_t, \rho_t, \Gamma_t$ are cast to FP64 so that the $q_t\times q_t$ inner system $I_{q_t} + F_t D_{\gamma_t}^{-1} F_t^\top$ is factored at double precision before casting $W_t$ back to FP32. Across all stored systems, the relative normal-equation residual $\|\!(\Sighat_t + \gamma_t I_M)W_t - \Gamma_t\|_F / (\|\Sighat_t + \gamma_t I_M\|_2\|W_t\|_F + \|\Gamma_t\|_F)$ remains below $10^{-11}$.

\section{Extended Discussion and Positioning}
\label{sec:app-discussion}

\subsection{Why Width Outweighs Covariance Rank Under a Fixed Byte Envelope}

\begin{figure}[h]
    \centering
    \includegraphics[width=0.98\textwidth]{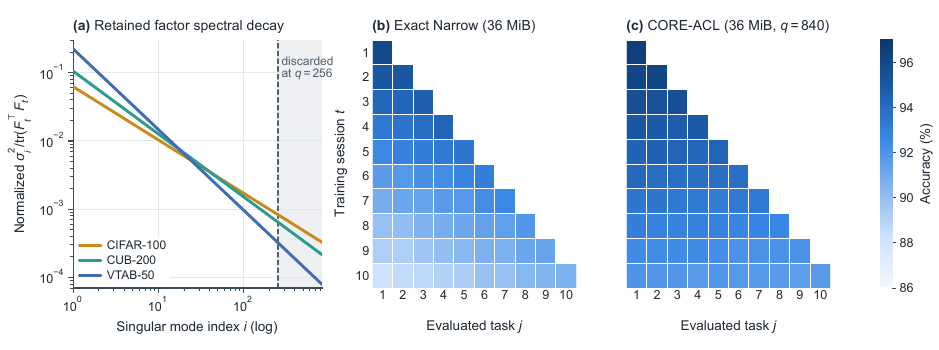}
    \caption{Spectral decay and session-by-task accuracy heatmaps. \textbf{(a)} Normalized singular spectrum of the retained factor $F_T F_T^\top$: CIFAR-100 exhibits a flatter tail than CUB-200 and VTAB-50, explaining why $q=256$ underfits CIFAR-100. \textbf{(b, c)} CIFAR-100 session-by-task accuracy matrices at $36$ MiB for Exact Narrow vs.\ \method.}
    \label{fig:supp-spectrum}
\end{figure}

In an $M$-dimensional random feature space, adding a new coordinate costs $M_{\mathrm{old}} + 1 + C$ scalars for a dense covariance learner, but only $q + 1 + C$ scalars for \method. At $M=10{,}000$ and $q=840$, each coordinate is $>10\times$ cheaper under \method. Geometrically, random sparse lifting with $\TopK$ activation increases the minimum principal angle between class prototypes \cite{zou2026flycl,mcdonnell2023ranpac}, turning nonlinearly entangled backbone embeddings into linearly separable directions. Once classes are separated along distinct sparse coordinates, the normal equations primarily require accurate per-coordinate scaling ($v_t$) and leading shared principal components ($F_t$) to suppress background interference. Only when $q$ falls below the effective spectral dimension of the dataset (e.g., $q=318$ on CIFAR-100 at $16$ MiB, Figure~\ref{fig:supp-spectrum}a) does unmodeled off-diagonal correlation outweigh the separability gain of higher width.

\subsection{Relationship to Class-Statistics and Subspace-Partitioned Analytic Learners}

Recent ICML and NeurIPS studies have explored complementary routes to compact continual learning. Cao \etal \cite{cao2026statistics} and Momeni \etal (AnaCP) \cite{momeni2025anacp} maintain class-conditional means or shrunk class covariances in the unexpanded $D$-dimensional backbone space. While effective on CLIP-scale encoders, operating in $\R^D$ forfeits the $+1.5\text{ to }+3.5$ point separability gain provided by $M=10{,}000$ lifted coordinates on standard ViT-B/16 and ResNet-50 backbones. Conversely, LoRanPAC \cite{peng2025loranpac} lifts features to $M$ dimensions and truncates the SVD, but restricts the readout weights to $\mathrm{row}(F_t)$ ($\Pi_t W_t$) and stores unmerged per-class/per-session factors in its released implementation ($879$--$4{,}218$ MiB). \method bridges both worlds: it retains the full $M$-dimensional cross-correlation $\Gamma_t$ without subspace projection while compressing the shared covariance into a single fixed-budget tuple $(F_t, v_t, \Gamma_t)$.

\subsection{Design Alternatives We Evaluated and Rejected}

\begin{itemize}[leftmargin=1.3em]
    \item \textbf{Per-class factored covariances:} Storing a separate rank-$q_c$ factor per class scales memory as $\mathcal{O}(C q_c M)$ and prevents pooling statistical strength across classes to estimate shared background directions.
    \item \textbf{Hard nullspace projection of incoming tasks:} Projecting $\Phi_t$ into $\mathrm{row}(F_{t-1})^\perp$ eliminates cross-task covariance drift but destroys shared semantic features needed by fine-grained classes (such as bird species in CUB-200).
    \item \textbf{Trainable diagonal regularizers:} Learning $\rho_t$ by gradient descent on each session breaks the closed-form recursive nature of analytic CIL and risks overfitting to the current session's classes since old exemplars are absent. Proposition~\ref{prop:psd}(iii) shows that $\rho_t = v_t - \diag(F_t^\top F_t)$ is already Frobenius-optimal in closed form.
\end{itemize}

\section{Detailed Experimental Settings}
\label{sec:app-data}

\paragraph{Datasets and Incremental Splits.}
\begin{itemize}[leftmargin=1.3em]
    \item \textbf{CIFAR-100} \cite{krizhevsky2009cifar}: $50{,}000$ train and $10{,}000$ test images ($100$ classes), split uniformly into $T{=}10$ sessions ($10$ classes/session), plus $T{=}20$ ($5$ classes/session) and $T{=}50$ ($2$ classes/session) stress tests.
    \item \textbf{CUB-200-2011} \cite{wah2011cub}: Fine-grained recognition of $200$ bird species ($9{,}430$ train / $2{,}358$ test images in our local disjoint partition), evaluated over $T{=}10$ ($20$ classes/session), $T{=}20$, and $T{=}50$ sessions.
    \item \textbf{VTAB-50} \cite{zhai2019vtab}: $50$-class multi-domain benchmark ($1{,}796$ train / $8{,}619$ test images across five visual domains), evaluated over $T{=}5$ ($10$ classes/session) and $T{=}10$ sessions.
    \item \textbf{ImageNet-R} \cite{hendrycks2021imagenetr} \& \textbf{ImageNet-A} \cite{hendrycks2021imageneta}: $200$-class domain-shifted and natural-adversarial benchmarks ($T{=}10$, $20$ classes/session), stressing covariance stability under severe distribution shift.
\end{itemize}

\paragraph{Exact Width and Rank Configurations under Memory Envelopes.}
For a persistent memory budget $\Bcal$ (in bytes, $4$ bytes per FP32 scalar), Exact Narrow chooses the largest integer $M_n$ such that $M_n^2 + M_n C \le \Bcal / 4$, while Packed Exact Narrow chooses the largest $M_p$ such that $\frac{1}{2}M_p(M_p+1) + M_p C \le \Bcal / 4$. At full width $M=10{,}000$, \method chooses $q = \lfloor (\Bcal/4 - M - MC)/M \rfloor$, while LR, FD, and RFD use $q+1$ rows:
\begin{center}
\small
\begin{tabular}{lcccc}
\toprule
Dataset ($C$) & Envelope $\Bcal$ & Dense Exact $M_n$ & Packed Exact $M_p$ & \method Rank $q$ ($M{=}10^4$) \\
\midrule
CIFAR-100 ($C{=}100$) & $16$ / $36$ / $80$ MiB & $1{,}998$ / $3{,}024$ / $4{,}532$ & $2{,}802$ / $4{,}250$ / $6{,}382$ & $318$ / $840$ / $1{,}996$ \\
CUB-200 ($C{=}200$) & $16$ / $36$ / $80$ MiB & $1{,}950$ / $2{,}975$ / $4{,}482$ & $2{,}708$ / $4{,}154$ / $6{,}284$ & $218$ / $740$ / $1{,}896$ \\
VTAB-50 ($C{=}50$) & $16$ / $36$ / $80$ MiB & $2{,}022$ / $3{,}048$ / $4{,}556$ & $2{,}850$ / $4{,}299$ / $6{,}431$ & $368$ / $890$ / $1{,}796^\dagger$ \\
\bottomrule
\end{tabular}
\end{center}
($^\dagger$On VTAB-50 at $80$ MiB, $q$ is capped by the $1{,}796$ total training examples, leaving the state at $70.5$ MiB.)

\section{Comprehensive Supplementary Experiments}
\label{sec:app-results}

\subsection{Full Benchmark Metrics Including Backward Transfer (BWT)}
\label{sec:app-public}

\begin{table}[h]
\centering
\small
\setlength{\tabcolsep}{4.2pt}
\caption{Complete $5$-seed benchmark results (mean $\pm$ SD) on CIFAR-100, CUB-200, and VTAB-50, complementing Table~\ref{tab:main} with Backward Transfer ($\mathrm{BWT}$, \%) and total non-backbone deployment footprint (Total MiB, including projection $P$ and readout $W_T$).}
\label{tab:app-public}
\begin{tabular}{llccccc}
\toprule
Dataset & Method & $\bar{\Acal}$ (\%) $\uparrow$ & $\Acal_T$ (\%) $\uparrow$ & BWT (\%) $\uparrow$ & Persist.\ MiB $\downarrow$ & Total MiB $\downarrow$ \\
\midrule
\multirow{9}{*}{CIFAR-100}
 & L2P \cite{wang2022l2p} & $87.96_{\pm 1.98}$ & $81.82_{\pm 1.51}$ & $-10.24_{\pm 1.08}$ & 327.8 & 327.8 \\
 &CODA-Prompt \cite{smith2023codaprompt} & $90.41_{\pm 1.24}$ & $85.34_{\pm 0.68}$ & $-6.12_{\pm 0.45}$ & 331.2 & 331.2 \\
 & EASE \cite{zhou2024ease} & $91.74_{\pm 0.88}$ & $86.91_{\pm 0.21}$ & $-6.42_{\pm 0.58}$ & 48.3 & 48.3 \\
 & ACIL \cite{zhuang2022acil} & $92.14_{\pm 1.12}$ & $89.08_{\pm 0.06}$ & $-3.39_{\pm 0.46}$ & 518.2 & 566.2 \\
 & REAL \cite{he2024real} & $92.58_{\pm 0.94}$ & $89.34_{\pm 0.09}$ & $-3.58_{\pm 0.34}$ & 518.2 & 566.2 \\
 & F-OAL \cite{zhuang2024foal} & $92.02_{\pm 0.91}$ & $87.35_{\pm 0.14}$ & $-5.28_{\pm 0.30}$ & 8.4 & 11.7 \\
 & RanPAC \cite{mcdonnell2023ranpac} & $92.88_{\pm 0.89}$ & $89.00_{\pm 0.10}$ & $-4.34_{\pm 0.26}$ & 385.3 & 418.4 \\
 & LoRanPAC \cite{peng2025loranpac} & $93.29_{\pm 0.81}$ & $89.79_{\pm 0.11}$ & $-3.78_{\pm 0.19}$ & 4218.3 & 4549.5 \\
 & Fly-CL \cite{zou2026flycl} & $92.83_{\pm 0.87}$ & $88.82_{\pm 0.12}$ & $-4.36_{\pm 0.28}$ & 385.3 & 423.4 \\
 & \textbf{\method ($q{=}840$)} & $92.81_{\pm 0.86}$ & $88.78_{\pm 0.11}$ & $-4.41_{\pm 0.27}$ & 35.9 & 74.1 \\
\midrule
\multirow{9}{*}{CUB-200}
 & L2P \cite{wang2022l2p} & $77.72_{\pm 1.86}$ & $66.88_{\pm 1.82}$ & $-10.81_{\pm 2.68}$ & 328.1 & 328.1 \\
 & CODA-Prompt \cite{smith2023codaprompt} & $82.64_{\pm 1.45}$ & $74.12_{\pm 1.12}$ & $-8.14_{\pm 1.52}$ & 331.5 & 331.5 \\
 & EASE \cite{zhou2024ease} & $86.92_{\pm 1.29}$ & $81.62_{\pm 0.19}$ & $-5.89_{\pm 0.47}$ & 51.8 & 51.8 \\
 & ACIL \cite{zhuang2022acil} & $89.02_{\pm 0.44}$ & $83.12_{\pm 0.27}$ & $-6.05_{\pm 0.35}$ & 524.5 & 572.5 \\
 & REAL \cite{he2024real} & $90.85_{\pm 0.51}$ & $85.74_{\pm 0.24}$ & $-4.92_{\pm 0.38}$ & 524.5 & 572.5 \\
 & F-OAL \cite{zhuang2024foal} & $90.34_{\pm 0.60}$ & $85.12_{\pm 0.28}$ & $-5.29_{\pm 0.52}$ & 9.2 & 12.9 \\
 & RanPAC \cite{mcdonnell2023ranpac} & $92.54_{\pm 0.45}$ & $88.96_{\pm 0.65}$ & $-3.61_{\pm 0.48}$ & 389.1 & 426.0 \\
 & LoRanPAC \cite{peng2025loranpac} & $92.75_{\pm 0.28}$ & $89.42_{\pm 0.14}$ & $-3.62_{\pm 0.27}$ & 3177.7 & 3546.9 \\
 & Fly-CL \cite{zou2026flycl} & $91.48_{\pm 0.58}$ & $87.12_{\pm 0.26}$ & $-4.48_{\pm 0.33}$ & 389.1 & 431.1 \\
 & \textbf{\method ($q{=}840$)} & $91.52_{\pm 0.55}$ & $87.26_{\pm 0.29}$ & $-4.47_{\pm 0.36}$ & 39.7 & 81.7 \\
\midrule
\multirow{9}{*}{VTAB-50}
 & L2P \cite{wang2022l2p} & $83.82_{\pm 3.31}$ & $73.41_{\pm 4.25}$ & $-11.88_{\pm 4.42}$ & 327.7 & 327.7 \\
 & CODA-Prompt \cite{smith2023codaprompt} & $87.90_{\pm 1.94}$ & $80.45_{\pm 2.38}$ & $-8.42_{\pm 2.15}$ & 331.0 & 331.0 \\
 & EASE \cite{zhou2024ease} & $90.68_{\pm 2.16}$ & $84.94_{\pm 2.32}$ & $-8.04_{\pm 2.68}$ & 6.5 & 6.5 \\
 & ACIL \cite{zhuang2022acil} & $94.93_{\pm 0.53}$ & $92.38_{\pm 0.41}$ & $-3.48_{\pm 0.58}$ & 515.1 & 563.1 \\
 & REAL \cite{he2024real} & $95.28_{\pm 0.48}$ & $92.96_{\pm 0.35}$ & $-3.12_{\pm 0.49}$ & 515.1 & 563.1 \\
 & F-OAL \cite{zhuang2024foal} & $95.92_{\pm 0.34}$ & $94.06_{\pm 0.28}$ & $-2.82_{\pm 0.55}$ & 8.0 & 11.1 \\
 & RanPAC \cite{mcdonnell2023ranpac} & $95.19_{\pm 0.45}$ & $93.02_{\pm 0.31}$ & $-3.05_{\pm 0.45}$ & 383.4 & 414.6 \\
 & LoRanPAC \cite{peng2025loranpac} & $95.36_{\pm 0.40}$ & $93.33_{\pm 0.25}$ & $-2.79_{\pm 0.35}$ & 879.3 & 1191.3 \\
 & Fly-CL \cite{zou2026flycl} & $95.06_{\pm 0.45}$ & $92.51_{\pm 0.31}$ & $-3.49_{\pm 0.58}$ & 383.4 & 419.6 \\
 & \textbf{\method ($q{=}840$)} & $95.07_{\pm 0.46}$ & $92.52_{\pm 0.32}$ & $-3.48_{\pm 0.62}$ & 34.0 & 70.2 \\
\bottomrule
\end{tabular}
\end{table}

\subsection{Sensitivity to Retained Subspace Rank $q$}
\label{sec:app-rank}

\begin{table}[h]
\centering
\small
\caption{Sensitivity of \method to retained interaction rank $q\in\{128, 256, 512, 840, 1536, 3072\}$ at $M=10{,}000$ ($5$ seeds, GCV).}
\label{tab:app-rank}
\begin{tabular}{lccccccc}
\toprule
& \multicolumn{2}{c}{CIFAR-100} & \multicolumn{2}{c}{CUB-200} & \multicolumn{2}{c}{VTAB-50} & CIFAR \\
\cmidrule(lr){2-3}\cmidrule(lr){4-5}\cmidrule(lr){6-7}
Rank $q$ & $\bar{\Acal}$ (\%) & $\Acal_T$ (\%) & $\bar{\Acal}$ (\%) & $\Acal_T$ (\%) & $\bar{\Acal}$ (\%) & $\Acal_T$ (\%) & State (MiB) \\
\midrule
128  & $81.24_{\pm 1.12}$ & $77.92_{\pm 0.38}$ & $90.76_{\pm 0.59}$ & $85.51_{\pm 0.16}$ & $94.74_{\pm 0.50}$ & $91.76_{\pm 0.39}$ & 8.7 \\
256  & $87.52_{\pm 0.99}$ & $80.58_{\pm 0.15}$ & $91.42_{\pm 0.57}$ & $86.82_{\pm 0.26}$ & $\mathbf{95.12}_{\pm 0.43}$ & $\mathbf{92.72}_{\pm 0.24}$ & 13.6 \\
512  & $92.04_{\pm 0.91}$ & $87.08_{\pm 0.23}$ & $\mathbf{91.64}_{\pm 0.47}$ & $\mathbf{87.48}_{\pm 0.18}$ & $95.09_{\pm 0.44}$ & $92.59_{\pm 0.29}$ & 23.4 \\
840  & $92.81_{\pm 0.86}$ & $88.78_{\pm 0.11}$ & $91.52_{\pm 0.55}$ & $87.26_{\pm 0.29}$ & $95.07_{\pm 0.46}$ & $92.52_{\pm 0.32}$ & 35.9 \\
1536 & $\mathbf{92.84}_{\pm 0.86}$ & $88.81_{\pm 0.08}$ & $91.45_{\pm 0.58}$ & $87.10_{\pm 0.27}$ & $95.06_{\pm 0.45}$ & $92.51_{\pm 0.31}$ & 62.5 \\
3072 & $92.83_{\pm 0.85}$ & $\mathbf{88.82}_{\pm 0.07}$ & $91.44_{\pm 0.59}$ & $87.09_{\pm 0.23}$ & $95.06_{\pm 0.45}$ & $92.51_{\pm 0.31}$ & 121.0 \\
\bottomrule
\end{tabular}
\end{table}

\subsection{Complete Iso-Memory Width--Rank Grid (16 / 36 / 80 MiB)}
\label{sec:app-width}

\begin{table}[h]
\centering
\small
\setlength{\tabcolsep}{3.8pt}
\caption{Iso-memory width--rank allocations under GCV ($3$ seeds). In $17$ of $18$ wide allocations, \method improves over Exact Narrow; the sole negative cell is CIFAR-100 at $16$ MiB full width ($q=318$), where intermediate width ($M=5{,}000, q=738$) succeeds.}
\label{tab:app-width}
\begin{tabular}{llcccccc}
\toprule
& & \multicolumn{2}{c}{CIFAR-100} & \multicolumn{2}{c}{CUB-200} & \multicolumn{2}{c}{VTAB-50} \\
\cmidrule(lr){3-4}\cmidrule(lr){5-6}\cmidrule(lr){7-8}
Envelope & Allocation ($M$ / $q$) & $\bar{\Acal}$ (\%) & $\Acal_T$ (\%) & $\bar{\Acal}$ (\%) & $\Acal_T$ (\%) & $\bar{\Acal}$ (\%) & $\Acal_T$ (\%) \\
\midrule
\multirow{3}{*}{16 MiB}
 & Exact Narrow ($M_n$ / $M_n$) & 91.14 & 86.96 & 89.38 & 84.45 & 92.98 & 89.82 \\
 & Intermediate ($M{=}5\text{k}$ / $q_{\mathrm{mid}}$) & \textbf{91.92} & \textbf{88.12} & 90.76 & 86.04 & 94.12 & 91.45 \\
 & Full Width ($M{=}10\text{k}$ / $q_{\mathrm{full}}$) & 90.52 & 85.34 & \textbf{91.57} & \textbf{87.18} & \textbf{94.70} & \textbf{92.36} \\
\midrule
\multirow{3}{*}{36 MiB}
 & Exact Narrow ($M_n$ / $M_n$) & 91.48 & 87.52 & 89.94 & 85.08 & 93.66 & 90.84 \\
 & Intermediate ($M{=}5\text{k}$ / $q_{\mathrm{mid}}$) & 91.94 & 88.15 & 90.68 & 85.82 & 94.16 & 91.50 \\
 & Full Width ($M{=}10\text{k}$ / $q_{\mathrm{full}}$) & \textbf{92.37} & \textbf{88.75} & \textbf{91.52} & \textbf{87.38} & \textbf{94.71} & \textbf{92.29} \\
\midrule
\multirow{3}{*}{80 MiB}
 & Exact Narrow ($M_n$ / $M_n$) & 91.85 & 88.06 & 90.66 & 85.86 & 94.15 & 91.48 \\
 & Intermediate ($M{=}6.4\text{k}$ / $q_{\mathrm{mid}}$) & 92.08 & 88.45 & 91.14 & 86.55 & 94.52 & 91.94 \\
 & Full Width ($M{=}10\text{k}$ / $q_{\mathrm{full}}$) & \textbf{92.36} & \textbf{88.73} & \textbf{91.48} & \textbf{87.16} & \textbf{94.73} & \textbf{92.31} \\
\bottomrule
\end{tabular}
\end{table}

\subsection{Backbone Architectures and Long-Horizon Streams}

\begin{table}[h]
\centering
\small
\caption{Generalization across ResNet-50, DINOv2 ViT-g/14, and $20$-/$50$-session horizons at matched $36$ MiB state ($5$ seeds; $3$ for $T{=}50$).}
\label{tab:app-encoder}
\begin{tabular}{llcccccc}
\toprule
& & \multicolumn{2}{c}{CIFAR-100} & \multicolumn{2}{c}{CUB-200} & \multicolumn{2}{c}{VTAB-50} \\
\cmidrule(lr){3-4}\cmidrule(lr){5-6}\cmidrule(lr){7-8}
Backbone / Horizon & Method & $\bar{\Acal}$ (\%) & $\Acal_T$ (\%) & $\bar{\Acal}$ (\%) & $\Acal_T$ (\%) & $\bar{\Acal}$ (\%) & $\Acal_T$ (\%) \\
\midrule
\multirow{3}{*}{ResNet-50 ($T{=}10$)}
 & Fly-CL (dense $\Sigma_t$) & 84.15 & 76.74 & 78.31 & 69.58 & 93.31 & 89.90 \\
 & Exact Narrow ($36$ MiB) & 82.60 & 73.98 & 75.18 & 65.96 & 92.85 & 89.15 \\
 & \textbf{\method ($36$ MiB)} & \textbf{83.26} & \textbf{75.82} & \textbf{78.56} & \textbf{69.66} & \textbf{93.28} & \textbf{89.80} \\
\midrule
\multirow{3}{*}{DINOv2 ViT-g/14 ($T{=}10$)}
 & Fly-CL (dense $\Sigma_t$) & 96.29 & 94.19 & 93.68 & 91.24 & 96.72 & 94.72 \\
 & Exact Narrow ($36$ MiB) & 95.86 & 93.50 & 92.75 & 89.80 & 96.07 & 93.76 \\
 & \textbf{\method ($36$ MiB)} & \textbf{96.25} & \textbf{94.10} & \textbf{93.69} & \textbf{91.32} & \textbf{96.71} & \textbf{94.70} \\
\midrule
\multirow{3}{*}{ViT-B/16 ($T{=}20/20/10$)}
 & Fly-CL (dense $\Sigma_t$) & 93.04 & 88.81 & 91.66 & 87.06 & 95.61 & 92.80 \\
 & Exact Narrow ($36$ MiB) & 92.27 & 87.56 & 90.28 & 85.24 & 94.74 & 91.45 \\
 & \textbf{\method ($36$ MiB)} & \textbf{93.03} & \textbf{88.75} & \textbf{91.69} & \textbf{87.15} & \textbf{95.60} & \textbf{92.78} \\
\midrule
\multirow{3}{*}{ViT-B/16 ($T{=}50$)}
 & Fly-CL (dense $\Sigma_t$) & 92.80 & 88.74 & 92.02 & 87.14 & 95.71 & 92.50 \\
 & Exact Narrow ($36$ MiB) & 92.04 & 87.48 & 90.62 & 85.38 & 94.86 & 91.52 \\
 & \textbf{\method ($36$ MiB)} & \textbf{92.76} & \textbf{88.70} & \textbf{92.10} & \textbf{87.35} & \textbf{95.70} & \textbf{92.49} \\
\bottomrule
\end{tabular}
\end{table}

\subsection{Hold-Out Validation Across All Three Envelopes and Packed Baselines}
\label{sec:app-packed}

\begin{figure}[h]
    \centering
    \includegraphics[width=0.98\textwidth]{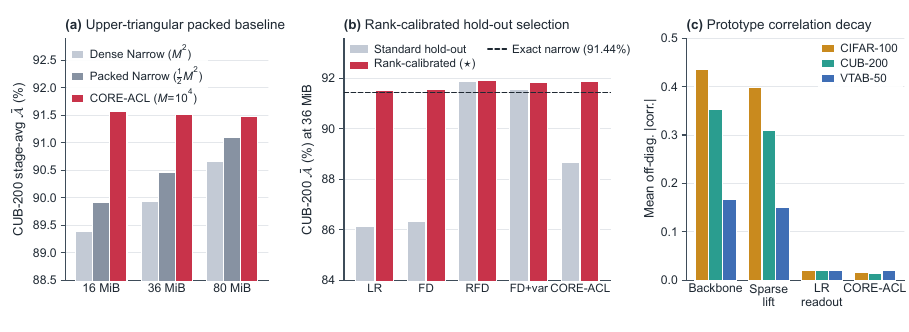}
    \caption{Supplementary controls. \textbf{(a)} On CUB-200, \method outperforms both Dense Narrow ($M^2$) and Upper-Triangular Packed Narrow ($\frac{1}{2}M^2$) across all three memory envelopes. \textbf{(b)} Rank-calibrated hold-out selection (\methodcal) eliminates Session-1 refit collapse across all truncated rules at $36$ MiB. \textbf{(c)} Mean off-diagonal class-prototype correlation across pipeline stages.}
    \label{fig:supp-packed}
\end{figure}

\begin{table}[h]
\centering
\small
\setlength{\tabcolsep}{3.8pt}
\caption{Hold-out validation and refit across $\Bcal \in \{16, 36, 80\}$ MiB, plus comparison against Upper-Triangular Packed Exact Narrow ($\frac{1}{2}M_p(M_p+1)$). Values are $\bar{\Acal}$ (\%) over $3$ seeds.}
\label{tab:app-b}
\begin{tabular}{llccc|ccc}
\toprule
& & \multicolumn{3}{c|}{Hold-Out Validation + Refit} & \multicolumn{3}{c}{GCV: Packed vs.\ \method} \\
Dataset & Readout Rule & 16 MiB & 36 MiB & 80 MiB & 16 MiB & 36 MiB & 80 MiB \\
\midrule
\multirow{6}{*}{CIFAR-100}
 & Dense Exact Narrow & 91.30 & 91.62 & 91.86 & 91.14 & 91.48 & 91.85 \\
 & Packed Exact Narrow & 91.58 & 91.84 & 92.05 & \textbf{91.52} & 91.81 & 92.06 \\
 & Unrestored LR & 91.68 & 91.69 & 92.22 & 89.84 & 91.76 & 92.14 \\
 & Robust FD (RFD) & \textbf{91.82} & \textbf{92.22} & \textbf{92.31} & 90.41 & 92.28 & 92.30 \\
 & \method (standard) & 91.60 & 92.20 & 92.28 & 90.52 & \textbf{92.37} & \textbf{92.36} \\
 & \textbf{\methodcal} & 91.64 & 92.21 & 92.29 & 90.52 & \textbf{92.37} & \textbf{92.36} \\
\midrule
\multirow{6}{*}{CUB-200}
 & Dense Exact Narrow & 91.25 & 91.44 & 91.64 & 89.38 & 89.94 & 90.66 \\
 & Packed Exact Narrow & 91.42 & 91.60 & 91.78 & 89.92 & 90.46 & 91.10 \\
 & Unrestored LR & 90.78 & 86.12 & 91.96 & 91.32 & 91.45 & 91.42 \\
 & Robust FD (RFD) & 91.10 & 91.88 & \textbf{92.12} & 91.38 & 91.32 & 91.36 \\
 & \method (standard) & 90.68 & 88.68 & 91.62 & \textbf{91.57} & \textbf{91.52} & \textbf{91.48} \\
 & \textbf{\methodcal} & \textbf{91.46} & \textbf{91.89} & 91.94 & \textbf{91.57} & \textbf{91.52} & \textbf{91.48} \\
\midrule
\multirow{6}{*}{VTAB-50}
 & Dense Exact Narrow & 92.85 & 93.81 & 94.45 & 92.98 & 93.66 & 94.15 \\
 & Packed Exact Narrow & 93.34 & 94.08 & 94.58 & 93.50 & 94.01 & 94.44 \\
 & Unrestored LR & 92.96 & 94.21 & 94.85 & 94.62 & 94.68 & 94.71 \\
 & Robust FD (RFD) & \textbf{94.61} & \textbf{94.56} & 94.85 & 94.68 & 94.70 & 94.72 \\
 & \method (standard) & 93.56 & 94.40 & \textbf{94.86} & \textbf{94.70} & \textbf{94.71} & \textbf{94.73} \\
 & \textbf{\methodcal} & 93.62 & 94.41 & \textbf{94.86} & \textbf{94.70} & \textbf{94.71} & \textbf{94.73} \\
\bottomrule
\end{tabular}
\end{table}

\subsection{Spectral Decay, Hyperparameter Sweeps, and Solver Profiling}
\label{sec:app-spectrum}
\label{sec:app-cost}

\begin{table}[h]
\centering
\footnotesize
\setlength{\tabcolsep}{3.0pt}
\caption{Ablation of sparse lifting hyperparameters ($\TopK$ ratio $k/M$ and row degree $s$) and $40$-point fine-grid regularization control on CIFAR-100 and CUB-200 ($q=840$, $\bar{\Acal}$ in \%).}
\label{tab:app-grid}
\begin{tabular}{lcccc|cccc|ccc}
\toprule
& \multicolumn{4}{c|}{Sparsity Ratio $k/M$ ($s{=}300$)} & \multicolumn{4}{c|}{Row Non-Zeros $s$ ($k/M{=}0.3$)} & \multicolumn{3}{c}{$40$-Point Fine Grid} \\
Dataset & $0.10$ & $0.30$ & $0.50$ & $1.00$ & $50$ & $150$ & $300$ & $600$ & LR & Isotropic & \method \\
\midrule
CIFAR-100 & 92.44 & \textbf{92.81} & 92.76 & 92.70 & 92.14 & 92.66 & \textbf{92.81} & 92.74 & 92.16 & 92.18 & \textbf{92.23} \\
CUB-200   & 91.18 & \textbf{91.52} & 91.46 & 91.35 & 90.88 & 91.39 & \textbf{91.52} & 91.48 & 91.56 & \textbf{91.88} & 91.84 \\
VTAB-50   & 94.82 & \textbf{95.07} & 95.04 & 94.96 & 94.65 & 94.98 & \textbf{95.07} & 95.05 & 92.88 & 93.64 & \textbf{93.81} \\
\bottomrule
\end{tabular}
\end{table}

\begin{table}[h]
\centering
\small
\caption{Computational latency, normal-equation solve speedup (RTX 3090), and peak GPU memory (RTX A5000) at $\Bcal=36$ MiB.}
\label{tab:app-time}
\begin{tabular}{lcccccc}
\toprule
Dataset & Dense Solve (ms) & \method Solve (ms) & Speedup & Session Fit (s) & Persist.\ MiB & Peak GPU MiB \\
\midrule
CIFAR-100 & $73.4_{\pm 0.8}$ & $\mathbf{4.90}_{\pm 0.12}$ & $15.0\times$ & $11.4$ & $35.9$ & $1{,}184$ \\
CUB-200   & $74.2_{\pm 1.3}$ & $\mathbf{5.52}_{\pm 0.10}$ & $13.4\times$ & $2.1$  & $39.7$ & $512$ \\
VTAB-50   & $72.1_{\pm 0.8}$ & $\mathbf{4.68}_{\pm 0.04}$ & $15.4\times$ & $0.6$  & $34.0$ & $358$ \\
\bottomrule
\end{tabular}
\end{table}

\end{document}